\pdfoutput=1
\documentclass[11pt]{article}

\usepackage[final]{acl}
\usepackage{arydshln}
\usepackage{times}
\usepackage{latexsym}

\usepackage[T1]{fontenc}
\usepackage[utf8]{inputenc}

\usepackage{microtype}

\usepackage{inconsolata}

\usepackage{tikz}
\usepackage{graphicx}
\usepackage{multirow}
\usepackage{amsmath}
\usepackage{amssymb}
\usepackage{wrapfig}   % 用于实现文字环绕图片
\usepackage{float}
\usepackage{algorithm}
\usepackage{algorithmic}
\usepackage{mdframed}
\usepackage{tcolorbox}
\usepackage{subcaption}
\usepackage{listings}
\usepackage{lipsum}
\usepackage{array}
\usepackage{chngcntr}
\usepackage{ragged2e}
\tcbuselibrary{breakable}
\usepackage{booktabs}
\usepackage{makecell}
\usepackage{caption}
\usepackage{xcolor} % Added for color definitions
\usepackage{pifont}  % For \ding{55} (a cross)

\definecolor{lightgray}{gray}{0.9} % 定义一个浅灰色
\usepackage{tcolorbox}
\tcbuselibrary{skins}
\usepackage[table,xcdraw]{xcolor} % 必须包含 table 选项
\usepackage{colortbl}

\usepackage{xstring}

\definecolor{processRed}{RGB}{178, 34, 34}   % 红色
\definecolor{goalGreen}{RGB}{34, 139, 34}    % 绿色

\newtcbox{\processbadge}{
  on line, 
  arc=0pt,                % 1是很小的圆角，0是直角
  colback=white, 
  colframe=processRed,
  boxrule=0.6pt,          % 变细，显得精致
  boxsep=0pt, left=2pt, right=2pt, top=1pt, bottom=1pt,
  fontupper=\sffamily\bfseries\scriptsize\color{processRed}
}

\newtcbox{\goalbadge}{
  on line, 
  arc=0pt,                % 1是很小的圆角
  colback=white, 
  colframe=goalGreen,
  boxrule=0.6pt,          % 变细
  boxsep=0pt, left=2pt, right=2pt, top=1pt, bottom=1pt,
  fontupper=\sffamily\bfseries\scriptsize\color{goalGreen}
}

\newtcbox{\mathWp}{
  on line, arc=1.5pt, colback=white, colframe=processRed,
  boxrule=0.6pt, boxsep=1pt, left=1pt, right=1pt, top=0.5pt, bottom=0.5pt,
  fontupper=\bfseries\color{processRed} % 注意：这里的 bfseries 对数学公式可能不生效，主要靠颜色
}

\newtcbox{\mathWg}{
  on line, arc=1.5pt, colback=white, colframe=goalGreen,
  boxrule=0.6pt, boxsep=1pt, left=1pt, right=1pt, top=0.5pt, bottom=0.5pt,
  fontupper=\bfseries\color{goalGreen}
}
\definecolor{myGreen}{RGB}{0, 150, 0}
\definecolor{myRed}{RGB}{200, 0, 0}
\definecolor{alfred_bg}{HTML}{FCE4D6} % 浅粉色
\definecolor{habitat_bg}{HTML}{E2EFDA} % 浅绿色

\definecolor{nav_bg}{HTML}{DAE8FC}   % 浅蓝色 (EB-Navigation)
\definecolor{manip_bg}{HTML}{E1D5E7} % 浅紫色 (EB-Manipulation)

\newcommand{\perf}[2]{%
    #1%
    \rlap{$
        \,_{\IfBeginWith{#2}{-}%
            {\color{myGreen}\text{\tiny{(#2)}}}%
            {\color{myRed}\text{\tiny{(#2)}}}%
        }
    $}%
}

\newtcolorbox{definitionbox}[1][]{%
  colback=blue!5,       % 背景浅蓝
  colframe=blue!50!black, % 边框颜色
  coltitle=white,       % 标题文字颜色
  colbacktitle=blue!60!black, % 标题背景色
  boxrule=1.5pt,                 % 边框粗细
  rounded corners,               % 设置为圆角
  fonttitle=\bfseries,  % 标题字体
  enhanced,
  attach boxed title to top left={yshift=-2mm,xshift=2mm},
  boxed title style={
    rounded corners,
    borderline west={0pt}{0pt}{white}, % 隐藏标题框自身的边框
    borderline east={0pt}{0pt}{white},
    borderline north={0pt}{0pt}{white},
    borderline south={0pt}{0pt}{white},
  },
  title=Definition,
  #1
}
\usepackage{hyperref}
\usepackage{fontawesome5} % 用于 GitHub 图标
\hypersetup{
    colorlinks=true,
    urlcolor=black % 或者你想要的颜色，如 midnightblue
}
\definecolor{MyBlue}{RGB}{60, 120, 220}

\title{Aligning Agentic World Models via Knowledgeable Experience Learning}

\author{
    Baochang Ren\textsuperscript{$\spadesuit$},
    Yunzhi Yao\textsuperscript{$\spadesuit$},
    Rui Sun\textsuperscript{$\diamondsuit$},
    Shuofei Qiao\textsuperscript{$\spadesuit$}, \\
    \textbf{Ningyu Zhang}\textsuperscript{$\spadesuit$}\footnotemark[1],
    \textbf{Huajun Chen}\textsuperscript{$\spadesuit$}\thanks{Corresponding author.}\\
    \textsuperscript{$\spadesuit$} Zhejiang University \quad
    \textsuperscript{$\diamondsuit$} University of California, Los Angeles \quad \\
    \texttt{\{baochang.ren, zhangningyu\}@zju.edu.cn} \\
    \vspace{-0.1em} \\ 
    \url{https://zjunlp.github.io/project/WorldMind/}
}

\begin{document}
\maketitle
\begin{abstract}
Current Large Language Models (LLMs) exhibit a critical modal disconnect: they possess vast semantic knowledge but lack the procedural grounding to respect immutable environmental constraints.
Consequently, while these agents implicitly function as world models, their simulations often suffer from \textit{physical hallucinations}—generating plans that are logically sound but physically unexecutable.
Existing alignment strategies predominantly rely on resource-intensive training or fine-tuning, which attempt to compress dynamic environmental rules into static model parameters.
However, such parametric encapsulation is inherently rigid, struggling to adapt to the open-ended variability of environmental dynamics without continuous, costly retraining.
To solve this fundamental misalignment between semantic planning and environmental reality without relying on parameter updates, we introduce \textbf{WorldMind}, a framework that autonomously constructs a symbolic World Knowledge Repository by synthesizing environmental feedback.
Specifically, it unifies Process Experience to enforce semantic physical feasibility via prediction errors and Goal Experience to guide task optimality through successful trajectories.
Experiments on EB-ALFRED and EB-Habitat demonstrate that WorldMind achieves the best performance compared to baselines, exhibiting remarkable cross-model and cross-environment transferability.
%, suggesting that decoupling world knowledge from model parameters establishes a scalable cognitive layer for robust embodied intelligence.
\end{abstract}

\begin{figure}[t!]
    \centering
    \resizebox{.45\textwidth}{!}{
    \includegraphics{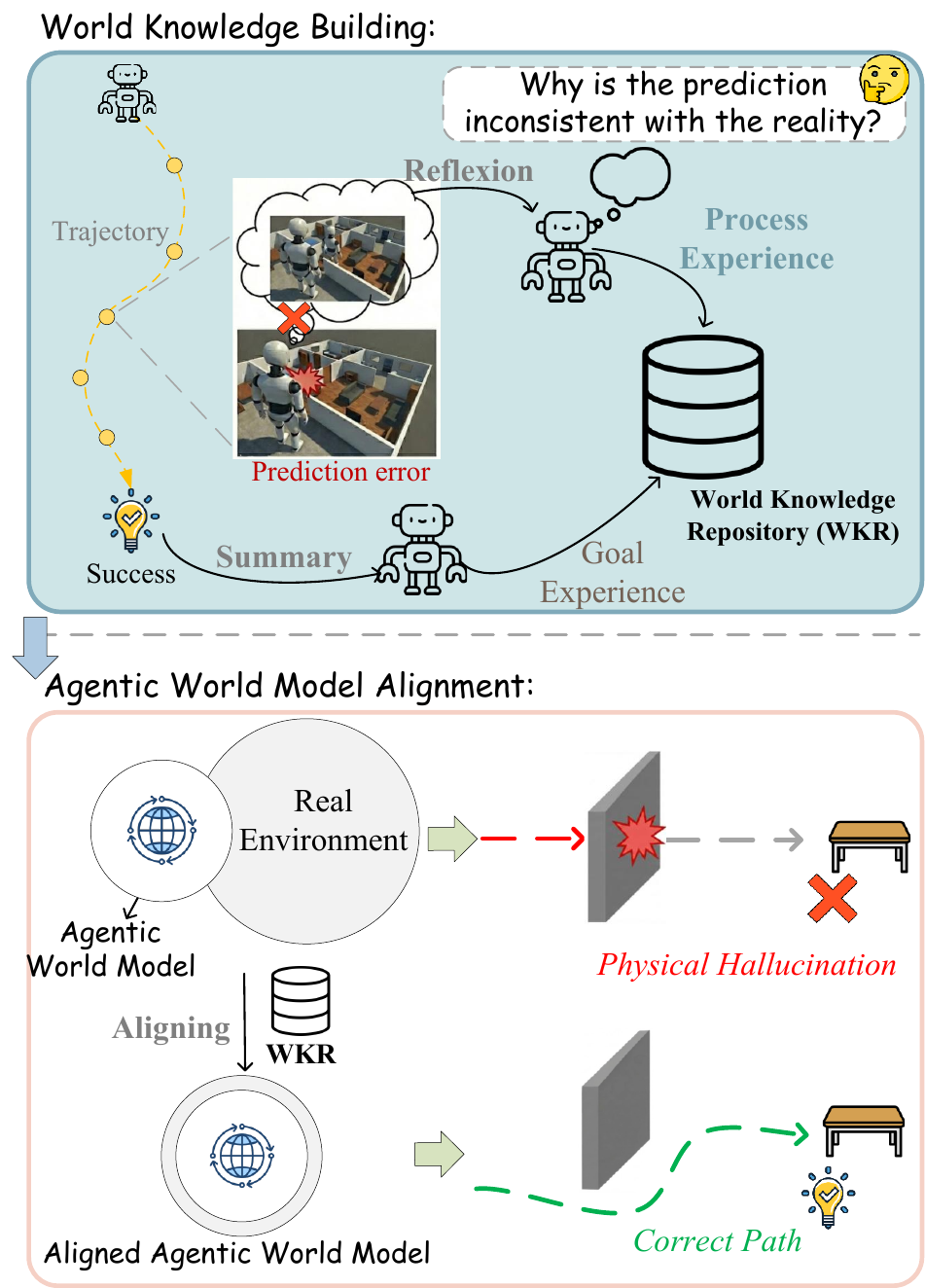}}
    \caption{Conceptual Illustration of Experiential Alignment. The agent aligns its internal world model via Process Experience and Goal Experience.}
     \vspace{-3ex}
    \label{fig:intro}
\end{figure}

\section{Introduction}

Building World Models capable of simulating and predicting environmental dynamics is foundational to the pursuit of embodied intelligence.
While traditional approaches rely on external simulators, a compelling paradigm has emerged where the agent itself functions as the world model~\citep{richens2025general}.
By internally encoding environmental dynamics, agents can reason about causal relationships and predict outcomes without heavy-weight external engines~\citep{zhang2021world, fang2025webevolver, chae2024web, wang2025adawm, yang2025mindjourney}.
Conceptually, this redefines the agent not merely as a passive policy executor, but as an active prediction machine of the world it inhabits~\citep{zhang2025agent, zheng2025flare, cen2025worldvla, wang2025world}.

However, a critical modal disconnect persists: while Large Language Models (LLMs) scale effectively on semantic reasoning, they exhibit a fundamental disparity between high-level logical planning and semantic physical constraints.
Despite possessing vast declarative knowledge, they often lack the procedural understanding required to respect the immutable laws of the physical world.
This misalignment leads to frequent physical hallucinations, where agents generate plans that are semantically coherent but physically unexecutable (e.g., attempting to slice an object without holding a knife).
The agent understands the high-level \textit{why}, but fails to grasp the procedural \textit{how} imposed by physical reality.

To bridge this gap, dominant paradigms currently rely on internalizing physical laws directly into model parameters via Supervised Fine-Tuning (SFT) or Reinforcement Learning~\citep{chae2024web, chen2025internalizing}.
We suggest that this approach may face inherent challenges, as it attempts to capture the variability of dynamic physical environments using static parameter weights.
Crucially, planning with an agentic world model presents a dual challenge: the internal simulation must not only identify the optimal path (Goal) but also strictly adhere to physical constraints (Process).
Existing agents often prioritize goal achievement while neglecting physical feasibility, leading to brittle plans that fail in dynamic settings.
Since static parameters cannot exhaustively cover all physical corner cases, an agent must possess the capacity for \textbf{online correction}, dynamically adjusting its understanding of constraints as it encounters them.
This context-based flexibility~\citep{zhang2025agentic, zhou2025memento, ouyang2025reasoningbank} raises a research question:
\textit{\textbf{Can agents autonomously align their world knowledge via experience learning in a training-free manner?}}

We answer this question by drawing inspiration from the cognitive theory of \textit{Predictive Coding}~\cite{huang2011predictive,friston2018does}.
In this view, intelligence is not the passive absorption of data, but the continuous minimization of ``prediction error''---the discrepancy between internal expectation and sensory reality.
Classic model-based reinforcement learning refines transition models from off-policy state--action pairs whose predicted next states diverge from observed outcomes.
We transplant this loop to training-free LLM agents by treating prediction-error episodes as off-policy evidence for a symbolic transition model, where a gradient update is replaced with a verbal causal-rule update.
Instead of discarding execution failures as noise, we leverage them as corrective signals.
When an agent's prediction contradicts reality, it exposes a boundary in the agent's internal world model.
We argue that LLM agents possess useful but miscalibrated predictive capabilities that can be aligned through these error signals during inference.

\begin{figure*}[t]
    \centering
    \includegraphics[width=1.0\linewidth]{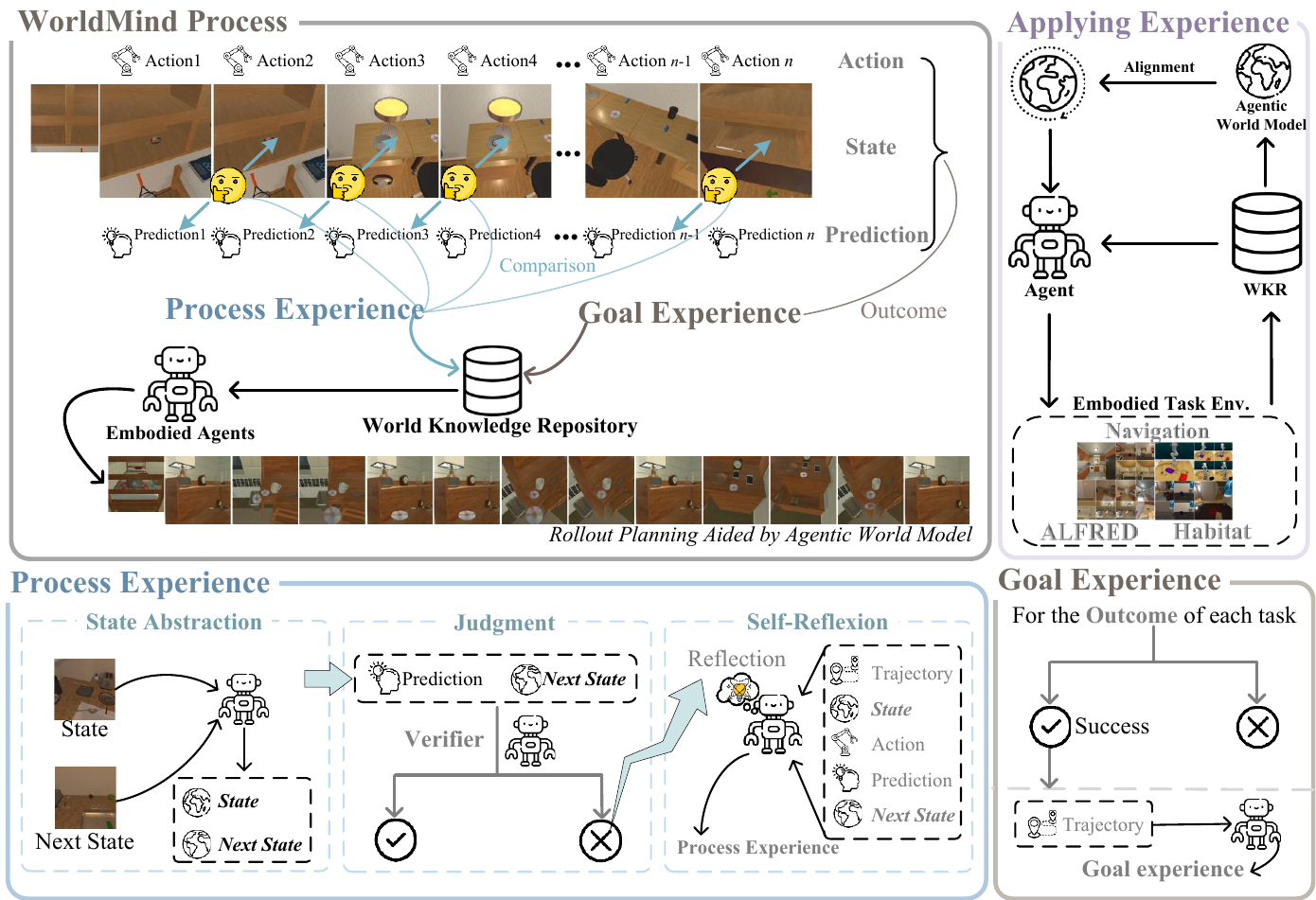}
    % \vspace{-2ex}
    \caption{\textbf{Overview of the WorldMind Framework.} The agent autonomously constructs a World Knowledge Repository (WKR) by unifying Process Experience (from prediction errors) and Goal Experience (from successful trajectories) to guide grounded simulation.}
    % \vspace{-3ex}
    \label{worldmind}
\end{figure*}

As shown in Figure~\ref{worldmind}, we propose \textbf{WorldMind}, a framework transforming autonomous agents into empirical world model learners.
To ensure robust simulations, our framework constructs a World Knowledge Repository comprising two distinct types of experience.
\textbf{Process Experience} leverages prediction errors to enforce physical feasibility, while \textbf{Goal Experience} distills successful trajectories into procedural heuristics to guide efficient task completion.
Extensive experiments on EB-ALFRED and EB-Habitat demonstrate that WorldMind improves over representative training-free baselines by reducing physically invalid actions and improving intermediate subgoal completion.
Controlled memory ablations further show that the gain is not simply due to adding a cross-task repository, and supplementary stress tests examine robustness under object renaming, memory corruption, and shuffled task order.
Cross-model and cross-environment results indicate that the constructed world knowledge transfers across model backbones and between household simulators, while we limit this claim to environments with overlapping semantics and physical assumptions.

\section{The WorldMind Framework}

\subsection{Overview and Problem Formulation}

As shown in Figure~\ref{worldmind}, we propose \textbf{WorldMind}, a framework transforming autonomous agents into empirical world model learners.
By externalizing environmental dynamics into explicit memory, we enable grounded planning through a cycle of goal alignment and continuous reality verification.

We formulate the problem as a \textbf{World Knowledge-Augmented Markov Decision Process (WK-MDP)}, defined by the tuple:
\begin{equation}
    \mathcal{M}_{\text{WK}} = \langle \rho_0 , \mathcal{S}, \mathcal{A}, \mathcal{P}, \Omega, \mathcal{G}, \mathcal{W} \rangle
\end{equation}
Here, $\mathcal{S}$ and $\mathcal{A}$ denote the state and action spaces, respectively. 
$\rho_0$ represents the initial state distribution from which the environment is reset at the beginning of each task episode.
$\mathcal{P}: \mathcal{S} \times \mathcal{A} \rightarrow \mathcal{S}$ represents the latent transition dynamics. 
The observation function $\Omega$ yields egocentric visual inputs $o_t$ based on the current state, and $\mathcal{G}$ represents the set of task objectives specified in natural language.

Distinct from standard MDPs, we introduce an explicit \textbf{World Knowledge Repository} $\mathcal{W} = \{ \mathcal{W}_p, \mathcal{W}_g \}$.
$\mathcal{W}_p$ encapsulates \processbadge{Process Experience}, a collection of verbalized causal rules derived from prediction errors.
This component functions as a learned surrogate model $\hat{\mathcal{P}}_p$ that progressively approximates the ground-truth dynamics $\mathcal{P}$, ensuring simulated steps adhere to physical constraints.
$\mathcal{W}_g$ encodes \goalbadge{Goal Experience}, a set of procedural heuristics distilled from successful trajectories.
These priors serve to constrain the policy search space towards the optimal policy $\pi^*$, ensuring the simulated path converges on the correct target.

The agent operates under a policy $\pi(a_t, \hat{s}_{t+1} \mid o_t, g, \mathcal{W})$ that jointly generates an action $a_t$ and a predicted future state $\hat{s}_{t+1}$ given a specific goal $g \in \mathcal{G}$.
The learning objective is formulated as:
\begin{equation}
    \max_{\pi} \mathbb{P}(\text{Success} \mid g)  \: \text{s.t.} \: \min \mathcal{D}(\hat{s}_{t+1} || s_{t+1})
\end{equation}
Here, $\text{Success}$ represents the evaluation metric directly provided by the simulation environment's underlying script, which automatically verifies whether the physical states of relevant objects satisfy the predefined conditions of the goal $g$.
This metric accounts for both final task completion and partial progress (such as completed subgoals).
This objective implies that maximizing the probability of task success is contingent upon minimizing the divergence $\mathcal{D}$ between the predicted state $\hat{s}$ and the actual state $s$.

\subsection{Building World Knowledge Repository}

The core of WorldMind lies in the autonomous development of the World Knowledge Repository (WKR) $\mathcal{W}$.
This process involves two distinct update mechanisms: one for establishing physical boundaries termed \processbadge{Process Experience}, and one for distilling procedural shortcuts known as \goalbadge{Goal Experience}.

\paragraph{Process Experience Construction.}
To align internal simulations with environmental reality, $\mathcal{M}$ operates in a \texttt{Predict-Act-Verify} loop.
At each step $t$, conditioned on the environmental prior $p_{env}$, current observation $o_t$, task goal $g$, and retrieved knowledge $\mathcal{W}$, the agent jointly generates an action and a textual prediction:
\begin{equation}
    (a_t, \hat{s}_{t+1}) \sim \mathcal{M}(o_t, p_{env}, g, \mathcal{W})
\end{equation}
After execution, the environment yields the actual grounded state $s_{t+1}$.
We then operationalize the acquisition of \processbadge{Process Experience} through three distinct steps: State Abstraction, Judgment, and Self-Reflexion.

First, during \textbf{State Abstraction}, to prioritize abstract causal variables over low-level details, we explicitly define a semantic abstraction process to convert the ground-truth states into high-level descriptions:
\begin{equation}
    \bar{s}_k = \mathcal{M}_{\text{abs}}(s_k), \quad k \in \{t, t+1\}
\end{equation}
Second, in the \textbf{Judgment} phase, the agent acts as a verifier, comparing its internal prediction $\hat{s}_{t+1}$ against the actual abstract outcome $\bar{s}_{t+1}$.
Specifically, we prompt a Large Language Model to act as a Logical Consistency Validator.
This validator conservatively evaluates whether there is a direct, factual contradiction between the predicted and actual states, deliberately ignoring irrelevant descriptions or unexecuted intents (detailed prompts are provided in Appendix~\ref{appendix:prompt}).
A learning signal is triggered only when such an explicit semantic discrepancy is detected, indicating a physical hallucination.
Finally, through \textbf{Self-Reflexion}, the agent addresses the identified conflict by triggering a reflective module $\mathcal{R}$.
This module operates by prompting a Large Language Model to act as a Root Cause Analyst, analyzing the error context alongside the interaction history $\tau_{t-1}$ to synthesize a corrective causal rule (detailed reflection prompts are provided in Appendix~\ref{appendix:prompt}).
The update rule for \processbadge{Process Experience} is formulated as:
\begin{equation}
    \mathcal{W}_{p} \leftarrow \mathcal{W}_{p} \cup \{ \mathcal{R}(\tau_{t-1}, a_t, \bar{s}_t, \hat{s}_{t+1}, \bar{s}_{t+1}) \}
\end{equation}
This three-step mechanism is designed to capture causal dynamics while filtering trivial environmental noise.
For example, if the agent predicts that it can place an object while still far from the intended receptacle but the environment reports an invalid placement, the resulting Process Experience records the precondition that the agent must first navigate to the destination receptacle before executing \texttt{put} actions.
Appendix~\ref{appendix:algorithm_examples} provides pseudocode and concrete Process/Goal Experience examples.

\begin{table*}[t]
\centering
\setlength{\tabcolsep}{2pt} 
\renewcommand{\arraystretch}{1.1}
\scriptsize

\resizebox{\textwidth}{!}{%
\begin{tabular}{l|l|l|cccccc|cccccc} 
\toprule

% --- 表头区域 ---
\multirow{2}{*}{\textbf{Backbone}} & \multirow{2}{*}{\textbf{Method}} & \multirow{2}{*}{\textbf{Memory Scope}} & 
\multicolumn{6}{c|}{\cellcolor{nav_bg}\textbf{Success Rate (SR)}} & 
\multicolumn{6}{c}{\cellcolor{manip_bg}\textbf{Goal Condition (GC)}} \\ 
\cmidrule(lr){4-9} \cmidrule(lr){10-15}

& & & \cellcolor{nav_bg}\textbf{Avg} & \textbf{Base} & \textbf{Common} & \textbf{Complex} & \textbf{Visual} & \textbf{Spatial} & 
    \cellcolor{manip_bg}\textbf{Avg} & \textbf{Base} & \textbf{Common} & \textbf{Complex} & \textbf{Visual} & \textbf{Spatial} \\ 
\midrule

% =============================================
% Part 1: Open-source and Proprietary Models
% =============================================
\multicolumn{15}{c}{\textit{Open-source and Proprietary Models}} \\
\midrule
GPT-4o & ReAct & Episodic & \cellcolor{nav_bg}56.8 & 64 & 54 & 68 & 46 & 52 & \cellcolor{manip_bg}65.6 & 74.0 & 60.3 & 74.0 & 58.3 & \textbf{61.3} \\
GPT-4o-mini & ReAct & Episodic & \cellcolor{nav_bg}28.8 & 34 & 28 & 36 & 24 & 22 & \cellcolor{manip_bg}37.8 & 47.8 & 35.3 & 43.5 & 33.3 & 29.0 \\
Claude-3.7-Sonnet & ReAct & Episodic & \cellcolor{nav_bg}\textbf{67.2} & 68 & \textbf{68} & 70 & \textbf{68} & \textbf{62} & \cellcolor{manip_bg}67.5 & 72.0 & 66.0 & \textbf{76.7} & \textbf{63.0} & 59.7 \\
Gemini-1.5-Pro & ReAct & Episodic & \cellcolor{nav_bg}63.2 & \textbf{70} & 64 & \textbf{72} & 58 & 52 & \cellcolor{manip_bg}\textbf{67.9} & \textbf{74.3} & \textbf{66.7} & 76.5 & 62.8 & 59.0 \\
Llama-3.2-90B-Vis & ReAct & Episodic & \cellcolor{nav_bg}35.2 & 38 & 34 & 44 & 28 & 32 & \cellcolor{manip_bg}40.3 & 43.7 & 37.3 & 49.2 & 35.3 & 36.0 \\
InternVL2.5-78B & ReAct & Episodic & \cellcolor{nav_bg}37.0 & 41 & 40 & 39 & 16 & 49 & \cellcolor{manip_bg}39.4 & 42.3 & 35.3 & 43.3 & 35.7 & 40.3 \\
\midrule

% =============================================
% Part 2: Method Comparison
% =============================================
\multicolumn{15}{c}{\textit{Method Comparison}} \\ 
\midrule
\multirow{7}{*}{GPT-3.5-turbo} 
  & ReAct & Episodic & \cellcolor{nav_bg}44.4 & 52 & \textbf{48} & 52 & 32 & 38 & \cellcolor{manip_bg}50.4 & 55.3 & \textbf{53.5} & 55.3 & 42.7 & 45.0 \\
  & BoN & Episodic & \cellcolor{nav_bg}42.8 & 46 & 42 & 50 & \textbf{42} & 34 & \cellcolor{manip_bg}50.4 & 54.2 & 46.5 & 56.5 & \textbf{52.0} & 42.8 \\
  & SimuRA & Episodic & \cellcolor{nav_bg}45.2 & 50 & 42 & 54 & 38 & 42 & \cellcolor{manip_bg}53.6 & 57.8 & 47.8 & 59.7 & 48.5 & \textbf{54.3} \\
  & ReasoningBank & Global & \cellcolor{nav_bg}41.6 & 50 & 36 & 44 & 36 & 42 & \cellcolor{manip_bg}47.6 & 57.5 & 41.5 & 47.0 & 44.2 & 48.0 \\
  & Synapse & Global & \cellcolor{nav_bg}38.8 & 38 & 46 & 40 & 36 & 34 & \cellcolor{manip_bg}43.6 & 42.5 & 51.3 & 42.7 & 42.0 & 39.7 \\
  & AWM & Global & \cellcolor{nav_bg}40.0 & 46 & 32 & 48 & 40 & 34 & \cellcolor{manip_bg}46.2 & 53.2 & 39.2 & 50.7 & 47.0 & 41.0 \\
  & \textbf{WorldMind} & Global & \cellcolor{nav_bg}\textbf{48.0} & \textbf{58} & \textbf{48} & \textbf{56} & 34 & \textbf{44} & \cellcolor{manip_bg}\textbf{54.1} & \textbf{63.0} & 52.7 & \textbf{61.0} & 41.7 & 52.0 \\
\midrule
\multirow{7}{*}{GPT-4.1-mini} 
  & ReAct & Episodic & \cellcolor{nav_bg}41.2 & 50 & 40 & 46 & 38 & 32 & \cellcolor{manip_bg}47.5 & 55.3 & 42.8 & 52.2 & 47.2 & 39.8 \\
  & BoN & Episodic & \cellcolor{nav_bg}44.4 & 46 & 44 & 50 & \textbf{42} & 40 & \cellcolor{manip_bg}49.5 & 50.8 & 48.3 & 54.7 & \textbf{48.7} & 45.0 \\
  & SimuRA & Episodic & \cellcolor{nav_bg}45.6 & \textbf{52} & 44 & \textbf{54} & 38 & 40 & \cellcolor{manip_bg}52.2 & \textbf{61.0} & 50.3 & 58.2 & 45.3 & 46.3 \\
  & ReasoningBank & Global & \cellcolor{nav_bg}38.0 & 42 & 36 & 42 & 34 & 36 & \cellcolor{manip_bg}42.6 & 46.7 & 38.8 & 45.8 & 41.5 & 40.3 \\
  & Synapse & Global & \cellcolor{nav_bg}37.2 & 40 & 32 & 44 & 36 & 34 & \cellcolor{manip_bg}42.2 & 41.2 & 37.5 & 49.5 & 41.3 & 41.7 \\
  & AWM & Global & \cellcolor{nav_bg}41.2 & 44 & 36 & 48 & 38 & 40 & \cellcolor{manip_bg}46.0 & 48.3 & 42.0 & 52.5 & 44.3 & 42.7 \\
  & \textbf{WorldMind} & Global & \cellcolor{nav_bg}\textbf{49.2} & 50 & \textbf{58} & \textbf{54} & \textbf{42} & \textbf{42} & \cellcolor{manip_bg}\textbf{55.7} & \textbf{61.0} & \textbf{61.0} & \textbf{58.8} & 48.0 & \textbf{49.7} \\
\bottomrule
\end{tabular}
}
\caption{\textbf{Main Results on the EB-ALFRED Benchmark.} The top block reports ReAct reference performance across backbones, while the lower block compares training-free methods under the same benchmark protocol. The Memory Scope column distinguishes Episodic methods without a cross-task global repository from Global methods with a cross-task global memory repository. For GPT-3.5-turbo experiments, GPT-4.1-mini handles visual perception, while GPT-3.5-turbo performs subsequent planning and WorldMind reasoning.}
\vspace{-3ex}
\label{tab:main_alfred}
\end{table*}

\paragraph{Goal Experience Construction.}
Efficient task completion necessitates heuristic guidance, complementing the physical feasibility ensured by Process Experience.
We address this by distilling procedural heuristics from successful episodes.
Given a successful trajectory $\tau^*$, the agent $\mathcal{M}$ acts as a task analysis expert to autonomously analyze the interaction history.
Specifically, the agent is prompted to evaluate the step-by-step task progress, identifying key actions that advanced the goal to extract a single, abstract, and reusable workflow while filtering out context-specific noise (detailed extraction prompts are provided in Appendix~\ref{appendix:prompt}).
The Goal Experience is updated as follows:
\begin{equation}
    \mathcal{W}_{g} \leftarrow \mathcal{W}_{g} \cup \mathcal{M}(\tau^*)
\end{equation}
These distilled priors serve as optimized meta-instructions, limiting the agent's policy search space towards the optimal subspace in future tasks.

\subsection{Inference via Constrained Simulation}

In the inference phase, the agent $\mathcal{M}$ leverages the constructed World Knowledge Repository $\mathcal{W}$ to guide decision-making through constrained simulation.
Specifically, the system dynamically retrieves the top-$2$ relevant world knowledge from the repository through semantic similarity\footnote{We utilize the \texttt{SentenceTransformer} library to compute text embeddings and cosine similarity for retrieval.} to the current task goal $g$.
Conditioned on this augmented context, the agent jointly generates the action $a_t$ and the predicted future state $\hat{s}_{t+1}$.
Crucially, to mitigate hallucinations, the generation of $\hat{s}_{t+1}$ is gated via prompt-based heuristic rules: the agent is explicitly instructed to only simulate outcomes when the target object is explicitly grounded in the current observation $o_t$ or the repository $\mathcal{W}$, otherwise executing the action and outputting a predefined skip string without updating the internal world model.
This selective strategy significantly enhances inference efficiency.
Unlike approaches relying on computationally intensive multiple sampling or heavy external world models, our framework minimizes latency by bypassing redundant simulations for ungrounded states.

\begin{table*}[t]
\centering
\setlength{\tabcolsep}{2pt} 
\renewcommand{\arraystretch}{1.1}
\scriptsize

\resizebox{\textwidth}{!}{%
\begin{tabular}{l|l|l|cccccc|cccccc} 
\toprule

% --- 表头区域 ---
\multirow{2}{*}{\textbf{Backbone}} & \multirow{2}{*}{\textbf{Method}} & \multirow{2}{*}{\textbf{Memory Scope}} & 
% 左侧 SR 区块 (使用 nav_bg)
\multicolumn{6}{c|}{\cellcolor{nav_bg}\textbf{Success Rate (SR)}} & 
% 右侧 GC 区块 (使用 manip_bg)
\multicolumn{6}{c}{\cellcolor{manip_bg}\textbf{Goal Condition (GC)}} \\ 
\cmidrule(lr){4-9} \cmidrule(lr){10-15}

% --- 子标题 ---
& & & \cellcolor{nav_bg}\textbf{Avg} & \textbf{Base} & \textbf{Common} & \textbf{Complex} & \textbf{Visual} & \textbf{Spatial} & 
    \cellcolor{manip_bg}\textbf{Avg} & \textbf{Base} & \textbf{Common} & \textbf{Complex} & \textbf{Visual} & \textbf{Spatial} \\ 
\midrule

% =============================================
% Part 1: Open-source and Proprietary Models
% =============================================
\multicolumn{15}{c}{\textit{Open-source and Proprietary Models}} \\
\midrule
GPT-4o & ReAct & Episodic & \cellcolor{nav_bg}58.0 & 86 & 44 & 56 & \textbf{68} & 36 & \cellcolor{manip_bg}70.4 & 90.7 & 56.0 & 68.0 & \textbf{75.2} & \textbf{62.1} \\
GPT-4o-mini & ReAct & Episodic & \cellcolor{nav_bg}36.4 & 74 & 22 & 32 & 22 & 32 & \cellcolor{manip_bg}48.6 & 77.5 & 32.5 & 42.0 & 33.1 & 57.8 \\
Claude-3.7-Sonnet & ReAct & Episodic & \cellcolor{nav_bg}\textbf{61.2} & 90 & \textbf{58} & \textbf{58} & 62 & \textbf{38} & \cellcolor{manip_bg}\textbf{72.3} & \textbf{97.5} & \textbf{68.5} & \textbf{79.5} & 72.0 & 43.8 \\
Gemini-1.5-Pro & ReAct & Episodic & \cellcolor{nav_bg}57.2 & 92 & 52 & 48 & 56 & \textbf{38} & \cellcolor{manip_bg}61.0 & 92.5 & 53.5 & 49.5 & 59.4 & 50.0 \\
Llama-3.2-90B-Vis & ReAct & Episodic & \cellcolor{nav_bg}45.6 & \textbf{94} & 24 & 50 & 32 & 28 & \cellcolor{manip_bg}55.9 & 94.5 & 32.5 & 53.0 & 39.7 & 59.6 \\
InternVL2.5-78B & ReAct & Episodic & \cellcolor{nav_bg}47.6 & 84 & 30 & \textbf{58} & 34 & 32 & \cellcolor{manip_bg}58.6 & 82.0 & 43.0 & 59.0 & 63.9 & 45.1 \\
\midrule

% =============================================
% Part 2: Method Comparison
% =============================================
\multicolumn{15}{c}{\textit{Method Comparison}} \\ 
\midrule
\multirow{7}{*}{GPT-3.5-turbo} 
  & ReAct & Episodic & \cellcolor{nav_bg}43.6 & 82 & 30 & 34 & 38 & 34 & \cellcolor{manip_bg}50.4 & 84.0 & 34.0 & 40.0 & 43.4 & 50.4 \\
  & BoN & Episodic & \cellcolor{nav_bg}43.6 & 74 & 34 & 36 & 40 & 34 & \cellcolor{manip_bg}50.8 & 75.5 & 35.5 & 43.0 & 46.9 & 53.3 \\
  & SimuRA & Episodic & \cellcolor{nav_bg}48.0 & 82 & 40 & 36 & \textbf{44} & \textbf{38} & \cellcolor{manip_bg}55.3 & 83.0 & \textbf{47.0} & 40.0 & 48.9 & 57.8 \\
  & ReasoningBank & Global & \cellcolor{nav_bg}46.4 & 76 & 40 & 40 & 42 & 34 & \cellcolor{manip_bg}53.9 & 78.5 & 41.5 & 44.0 & 46.2 & \textbf{59.5} \\
  & Synapse & Global & \cellcolor{nav_bg}48.4 & 84 & 40 & 38 & 42 & \textbf{38} & \cellcolor{manip_bg}54.2 & 85.0 & 43.0 & 43.0 & 46.3 & 53.5 \\
  & AWM & Global & \cellcolor{nav_bg}46.4 & 78 & \textbf{42} & \textbf{44} & 36 & 32 & \cellcolor{manip_bg}50.9 & 78.0 & 43.0 & 48.0 & 39.7 & 45.7 \\
  & \textbf{WorldMind} & Global & \cellcolor{nav_bg}\textbf{48.8} & \textbf{86} & \textbf{42} & \textbf{44} & 40 & 32 & \cellcolor{manip_bg}\textbf{56.7} & \textbf{89.0} & \textbf{47.0} & \textbf{51.0} & \textbf{49.8} & 46.8 \\
\midrule
\multirow{7}{*}{GPT-4.1-mini} 
  & ReAct & Episodic & \cellcolor{nav_bg}41.6 & 74 & 26 & 32 & 40 & \textbf{36} & \cellcolor{manip_bg}47.4 & 76.0 & 30.0 & 37.0 & 43.7 & 50.3 \\
  & BoN & Episodic & \cellcolor{nav_bg}44.4 & 74 & 32 & 36 & 46 & 34 & \cellcolor{manip_bg}50.9 & 75.8 & 38.0 & 41.5 & 48.9 & 50.5 \\
  & SimuRA & Episodic & \cellcolor{nav_bg}48.4 & 84 & 42 & 40 & 44 & 32 & \cellcolor{manip_bg}55.4 & 85.0 & 45.5 & 47.0 & 48.2 & 51.5 \\
  & ReasoningBank & Global & \cellcolor{nav_bg}46.4 & 76 & 42 & 30 & \textbf{52} & 32 & \cellcolor{manip_bg}53.7 & 77.0 & 45.5 & 34.5 & \textbf{58.4} & \textbf{53.3} \\
  & Synapse & Global & \cellcolor{nav_bg}45.6 & 78 & 42 & 38 & 34 & \textbf{36} & \cellcolor{manip_bg}52.0 & 78.3 & 46.5 & 45.5 & 39.7 & 50.0 \\
  & AWM & Global & \cellcolor{nav_bg}43.6 & 74 & 32 & 38 & 44 & 30 & \cellcolor{manip_bg}50.8 & 74.8 & 37.0 & 44.5 & 47.9 & 49.7 \\
  & \textbf{WorldMind} & Global & \cellcolor{nav_bg}\textbf{50.8} & \textbf{86} & \textbf{44} & \textbf{44} & 46 & 34 & \cellcolor{manip_bg}\textbf{57.2} & \textbf{87.0} & \textbf{48.5} & \textbf{51.5} & 52.3 & 46.7 \\
\bottomrule
\end{tabular}
}
\caption{\textbf{Main Results on the EB-Habitat Benchmark.} The top block reports ReAct reference performance across backbones, while the lower block compares training-free methods under the same benchmark protocol. The Memory Scope column distinguishes Episodic methods without a cross-task global repository from Global methods with a cross-task global memory repository. Performance is evaluated on five subsets: Base, Common, Complex, Visual, and Spatial; best results within each model group are in \textbf{bold}.}

\label{tab:main_habitat}
\end{table*}

\section{Experiment}

\subsection{Experimental Settings.}

\paragraph{Datasets and Metrics.}
We evaluate our framework on the EB-ALFRED and EB-Habitat benchmarks from EmbodiedBench \citep{yang2025embodiedbench}, focusing on five fine-grained subsets: Base, Common Sense, Complex Instruction, Spatial Awareness, and Visual Appearance.
To quantify performance, we employ two complementary metrics: \textit{Success Rate (SR)} and \textit{Goal-Conditioned Success (GC)}.
SR is a strict binary metric (0 or 1) indicating whether the final goal is fully achieved.
In contrast, GC evaluates process adherence by awarding partial credit for completed subgoals, even if the episode ultimately fails.

\paragraph{Models and Baselines.}
We employ GPT-4.1-mini and GPT-3.5-turbo as the primary backbones for our WorldMind framework.
To evaluate effectiveness, using reported results of open- and closed-source models from EmbodiedBench, we compare against representative baselines, including Best-of-N (BoN), ReAct \citep{yao2022react}, Synapse \citep{zheng2023synapse}, SimuRA \citep{deng2025simura}, ReasoningBank \citep{ouyang2025reasoningbank}, and AWM \citep{wang2024agent}.
All methods are evaluated under the same EmbodiedBench observation interface, action space, episode split, and SR/GC metrics; the compared variable is the planning, memory, or verification mechanism introduced by each method.
To make this comparison explicit, Tables~\ref{tab:main_alfred} and~\ref{tab:main_habitat} include a Memory Scope column that distinguishes \textit{Episodic} methods, which do not maintain a cross-task global repository, from \textit{Global} methods.
Appendix~\ref{appendix:baseline_descriptions} gives concise baseline descriptions.

\subsection{Main Results}

% \paragraph{WorldMind consistently outperforms baselines in strict task completion.}
% As detailed in Table~\ref{tab:main_alfred} and Table~\ref{tab:main_habitat}, our framework achieves the highest performance on the strict Success Rate (SR) metric across both benchmarks.
% Specifically, using the GPT-4.1-mini backbone in the EB-Habitat environment, WorldMind reaches an SR of 50.8\%, surpassing the ReAct baseline (41.6\%) by a significant margin of 9.2\%.
% Similarly, on EB-ALFRED with GPT-3.5-turbo, our method improves the SR from 44.4\% to 48.0\%.
% Since SR is a binary metric that only counts fully successful episodes, these results empirically validate that WorldMind is more effective at overcoming physical dead-ends to reach the final goal state.
\paragraph{WorldMind consistently outperforms baselines in strict task completion.}
As shown in Table~\ref{tab:main_alfred} and Table~\ref{tab:main_habitat}, WorldMind achieves the highest SR among evaluated methods across both benchmarks.
On EB-Habitat with the GPT-4.1-mini backbone, it attains an SR of 50.8\%, outperforming ReAct by 9.2\%, while on EB-ALFRED with GPT-3.5-turbo, SR improves from 44.4\% to 48.0\%.
Given that SR strictly measures complete task success, these gains demonstrate WorldMind's enhanced ability to overcome physical dead-ends and reach final goal states more consistently in practice.

\paragraph{WorldMind achieves improved procedural correctness.}
As shown in Table~\ref{tab:main_alfred} and Table~\ref{tab:main_habitat}, beyond final outcomes, WorldMind outperforms baselines on the GC metric, which evaluates the correct execution of intermediate steps.
On EB-ALFRED with GPT-3.5-turbo, the average GC improves from 50.4\% to 54.1\%.
Simultaneously, on EB-Habitat with the GPT-4.1-mini backbone, the GC reaches 57.2\%, surpassing the ReAct baseline (47.4\%) by a large margin.
These gains indicate that even when full task completion fails, WorldMind executes a higher proportion of valid subgoals, demonstrating that the distilled Goal Experience effectively guides sequential task progression.

\paragraph{Robustness across diverse capabilities and backbones.}
Our framework exhibits consistent stability across fine-grained dimensions, particularly in challenging subsets like Visual Appearance and Spatial Awareness.
For instance, the EB-Habitat Base Success Rate reaches 86\% for both GPT-3.5-turbo and GPT-4.1-mini, a model-agnostic benefit further validated across leading proprietary and open-source models (see Appendix~\ref{appendix:proprietary_opensource}).
This stability stems from the Predict-Act-Verify loop filtering ungrounded hallucinations across different model scales.
Furthermore, the framework maintains a scalable, bounded repository footprint (see Appendix~\ref{appendix:memory_budget}) with negligible real-time retrieval latency (see Appendix~\ref{appendix:practical_overhead}).
In particular, constructing Process and Goal Experience incurs one-time token costs, while top-2 retrieval adds only 0.0028 seconds per step on average.
We also validate generalization across action granularity on EB-Navigation (see Appendix~\ref{appendix:nav_experiment}), additional model backbones (see Appendix~\ref{appendix:proprietary_opensource}), controlled-memory effects in Section~4.4, and adversarial, task-order, and failure-taxonomy robustness (see Appendix~\ref{appendix:robustness_analyses}).

\subsection{Ablation Study}

Table~\ref{tab:ablation} presents the ablation results isolating the specific impact of each component.
We observe that Goal Experience primarily enhances the Goal Condition (GC) metric, particularly on EB-ALFRED with GPT-3.5-turbo.
This indicates that high-level strategies effectively guide the agent through correct intermediate subgoals, preventing navigational drift.
In contrast, Process Experience drives a larger improvement in the strict Success Rate (SR).
By encoding semantic environmental constraints, this component acts as a safeguard against fatal execution errors, ensuring that planned actions are not just logically sound but also executable within the simulation.
Across both benchmarks, the full model generally achieves the best overall performance, suggesting that \textbf{combining goal-based heuristics with semantic physical verification yields complementary benefits for effective alignment.}

\begin{figure*}[t]
    \centering
    % --- 子图 1: 经验迁移实验 ---
    \begin{subfigure}[b]{0.34\linewidth}
        \centering
        \includegraphics[width=\linewidth]{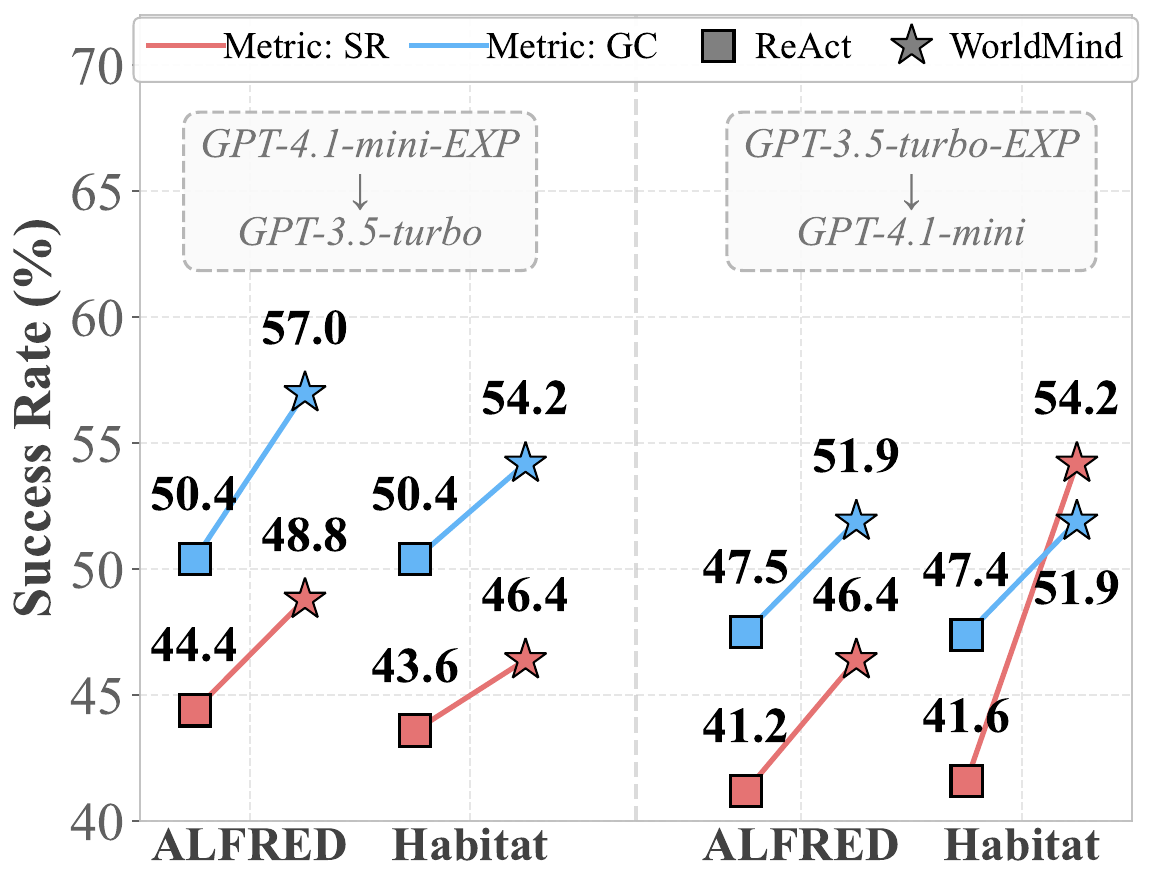} 
        \caption{Performance of Cross-Model Experience Transfer}
        \label{fig:transfer_exp}
    \end{subfigure}
    \hfill
    % --- 子图 2: Embodied Web Agent 准确率分析 (两个模型) ---
    \begin{subfigure}[b]{0.30\linewidth}
        \centering
        \includegraphics[width=\linewidth]{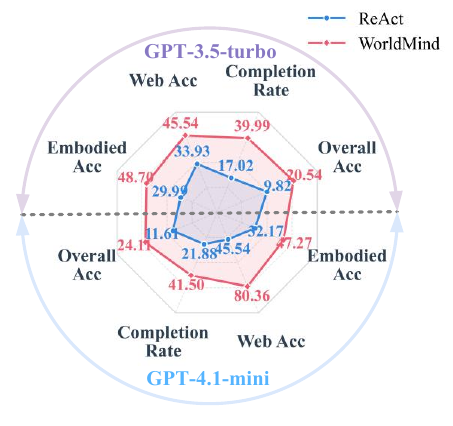}
        \caption{Accuracy Analysis on Embodied Web Agent Benchmark}
        \label{fig:web_accuracy}
    \end{subfigure}
    \hfill
    % --- 子图 3: Embodied Web Agent 错误分析 (两个模型) ---
    \begin{subfigure}[b]{0.34\linewidth}
        \centering
        \includegraphics[width=\linewidth]{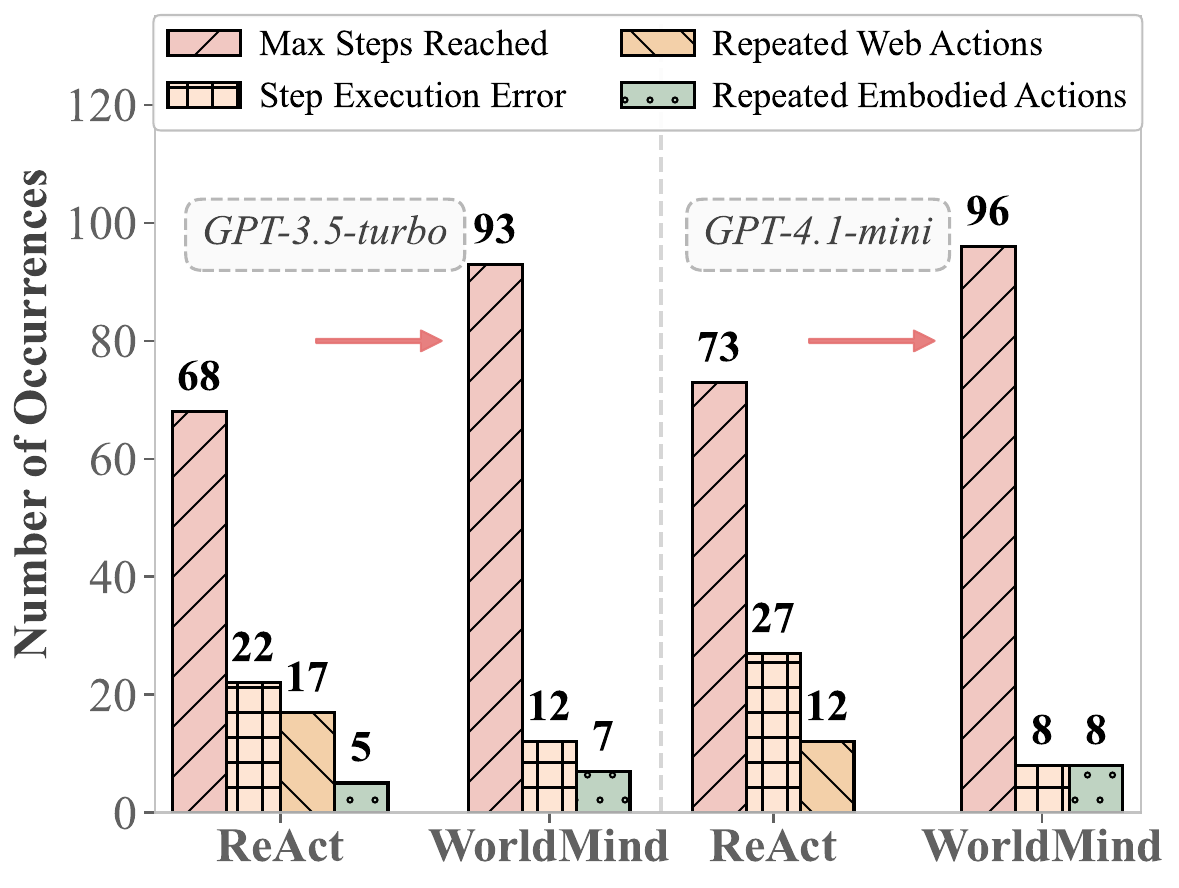}
        \caption{Error Analysis on Embodied Web Agent Benchmark}
        \label{fig:web_error}
    \end{subfigure}

    % --- 总标题 ---
    \caption{
        \textbf{Experimental Results and Analysis.} 
        (\subref{fig:transfer_exp}) Comparison of experience transfer capabilities between the two models. 
        (\subref{fig:web_accuracy}) Performance comparison on the Embodied Web Agent task, reporting accuracy metrics for both GPT-3.5-turbo and GPT-4.1-mini.
        (\subref{fig:web_error}) Comparative error distribution analysis for both models in the same environment.
    }
    \label{fig:main_results}
\end{figure*}

\section{Analysis}

\subsection{Cross-Model Experience Transfer Analysis}

Externalized symbolic knowledge should be reusable beyond the model that collected it.
To test this property, we exchange repositories between GPT-3.5-turbo and GPT-4.1-mini.
As shown in Figure~\ref{fig:transfer_exp}, GPT-3.5-turbo using GPT-4.1-mini's repository improves on EB-ALFRED (SR: 44.4\%$\to$48.8\%, GC: 50.4\%$\to$57.0\%), while GPT-4.1-mini using GPT-3.5-turbo's experience raises EB-Habitat SR from 41.6\% to 54.2\%.
\textbf{These bidirectional gains indicate that WorldMind captures transferable environmental regularities rather than model-specific traces.}

% \begin{table}[t]
% \centering
% \setlength{\tabcolsep}{8pt} % 调整列宽
% \renewcommand{\arraystretch}{1.2}

% \resizebox{\columnwidth}{!}{%
% \begin{tabular}{l|cc|c|c}
% \toprule
% \multirow{2}{*}{\textbf{Backbone}} & \multicolumn{2}{c|}{\textbf{Components}} & \textbf{EB-ALFRED} & \textbf{EB-Habitat} \\
% \cmidrule(lr){2-3} \cmidrule(lr){4-4} \cmidrule(lr){5-5}
% & \textbf{Goal} & \textbf{Process} & \textbf{Avg} & \textbf{Avg} \\
% \midrule

% % --- Group 1: GPT-3.5-turbo ---
% \multirow{4}{*}{GPT-3.5-turbo} 
%   & - & - & 44.4 & 43.6 \\ % Baseline (ReAct)
%   & \checkmark & - & - & - \\ % Only Goal (请填入数据)
%   & - & \checkmark & - & - \\ % Only Process (请填入数据)
%   & \checkmark & \checkmark & \textbf{48.0} & \textbf{48.8} \\ % WorldMind (Full)

% \midrule

% % --- Group 2: GPT-4.1-mini ---
% \multirow{4}{*}{GPT-4.1-mini} 
%   & - & - & 41.2 & 41.6 \\ % Baseline (ReAct)
%   & \checkmark & - & 48.8 & 48.4 \\ % Only Goal (请填入数据)
%   & - & \checkmark & 46.4 & 47.6 \\ % Only Process (请填入数据)
%   & \checkmark & \checkmark & \textbf{49.2} & \textbf{50.8} \\ % WorldMind (Full)

% \bottomrule
% \end{tabular}
% }
% \caption{Ablation study on the effectiveness of Goal Guidance and Process Supervision. The results report the average success rate on EB-ALFRED and EB-Habitat.}
% \label{tab:ablation}
% \end{table}

\begin{table}[h]
\centering
\setlength{\tabcolsep}{5pt} % 稍微减小列间距以容纳新增列
\renewcommand{\arraystretch}{1.2}

\resizebox{\columnwidth}{!}{%
\begin{tabular}{l|cc|cc|cc}
\toprule
\multirow{2}{*}{\textbf{Backbone}} & \multicolumn{2}{c|}{\textbf{Components}} & \multicolumn{2}{c|}{\textbf{EB-ALFRED}} & \multicolumn{2}{c}{\textbf{EB-Habitat}} \\
\cmidrule(lr){2-3} \cmidrule(lr){4-5} \cmidrule(lr){6-7}
& \textbf{Goal} & \textbf{Process} & \textbf{SR} & \textbf{GC} & \textbf{SR} & \textbf{GC} \\
\midrule

% --- Group 1: GPT-3.5-turbo ---
\multirow{4}{*}{GPT-3.5-turbo} 
  & - & - & 44.4 & 50.4 & 43.6 & 50.4 \\ % Baseline (ReAct)
  & \checkmark & - & 44.8 & 51.0 & 47.2 & 54.6 \\ % Only Goal
  & - & \checkmark & 42.4 & 47.5 & 45.2 & 51.3 \\ % Only Process
  & \checkmark & \checkmark & \textbf{48.0} & \textbf{54.1} & \textbf{48.8} & \textbf{56.7} \\ % WorldMind (Full)

\midrule

% --- Group 2: GPT-4.1-mini ---
\multirow{4}{*}{GPT-4.1-mini} 
  & - & - & 41.2 & 47.5 & 41.6 & 47.4 \\ % Baseline (ReAct)
  & \checkmark & - & 48.8 & \textbf{56.3} & 48.4 & 55.7 \\ % Only Goal
  & - & \checkmark & 46.4 & 51.5 & 47.6 & 54.0 \\ % Only Process
  & \checkmark & \checkmark & \textbf{49.2} & 55.7 & \textbf{50.8} & \textbf{57.2} \\ % WorldMind (Full)

\bottomrule
\end{tabular}
}
\caption{Ablation study on the effectiveness of Goal Guidance and Process Supervision. We report both SR and GC on EB-ALFRED and EB-Habitat.}

\label{tab:ablation}
\end{table}

\subsection{Cross-Environment Analysis}

To evaluate scalability, we further test on Embodied Web Agent \citep{hong2025embodied}, where agents switch between web information seeking and embodied procedural execution.
We use 112 sampled Indoor Cooking tasks and report Embodied Accuracy, Web Accuracy, Overall Accuracy, and Completion Rate.

\begin{figure}[h]
    \centering
    % --- 第一行 ---
    % 左图
    \begin{subfigure}[b]{0.48\linewidth} 
        \centering
        \includegraphics[width=\linewidth]{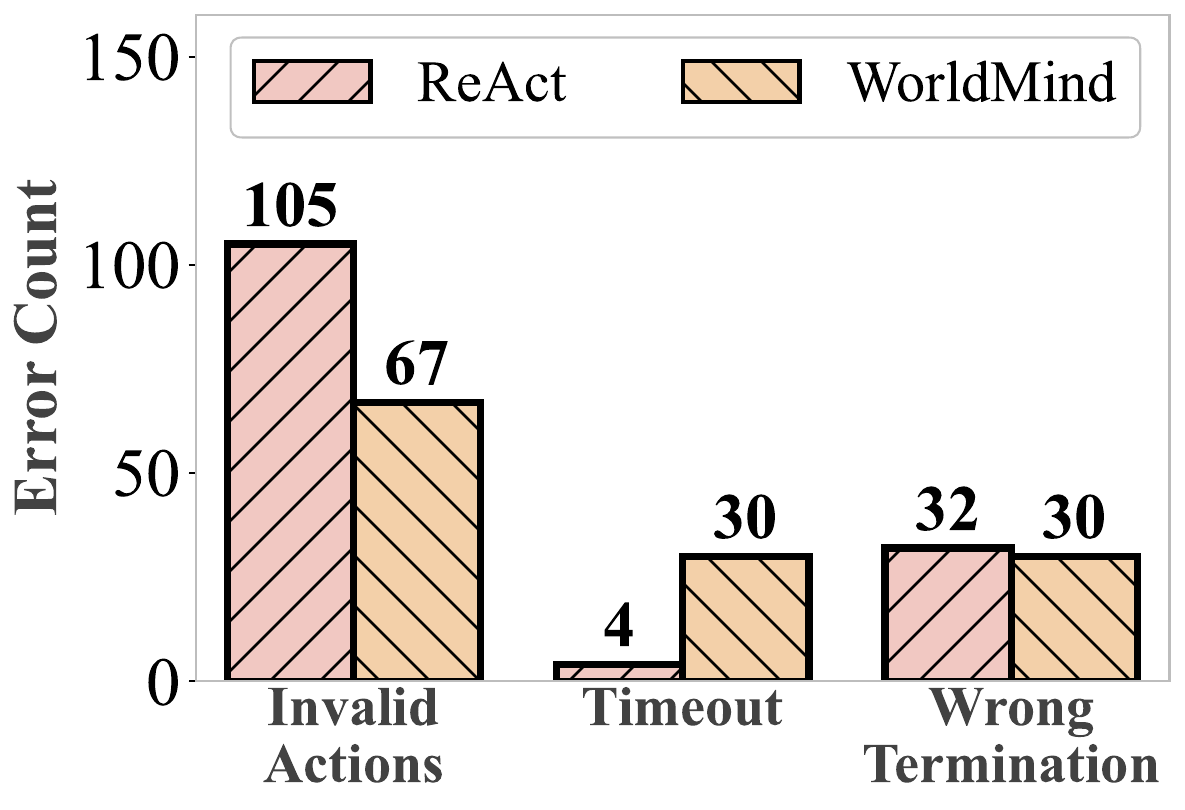}
        \caption{Habitat (GPT-3.5-turbo)}
        \label{fig:case1}
    \end{subfigure}
    \hfill % 左右间距
    % 右图
    \begin{subfigure}[b]{0.48\linewidth}
        \centering
        \includegraphics[width=\linewidth]{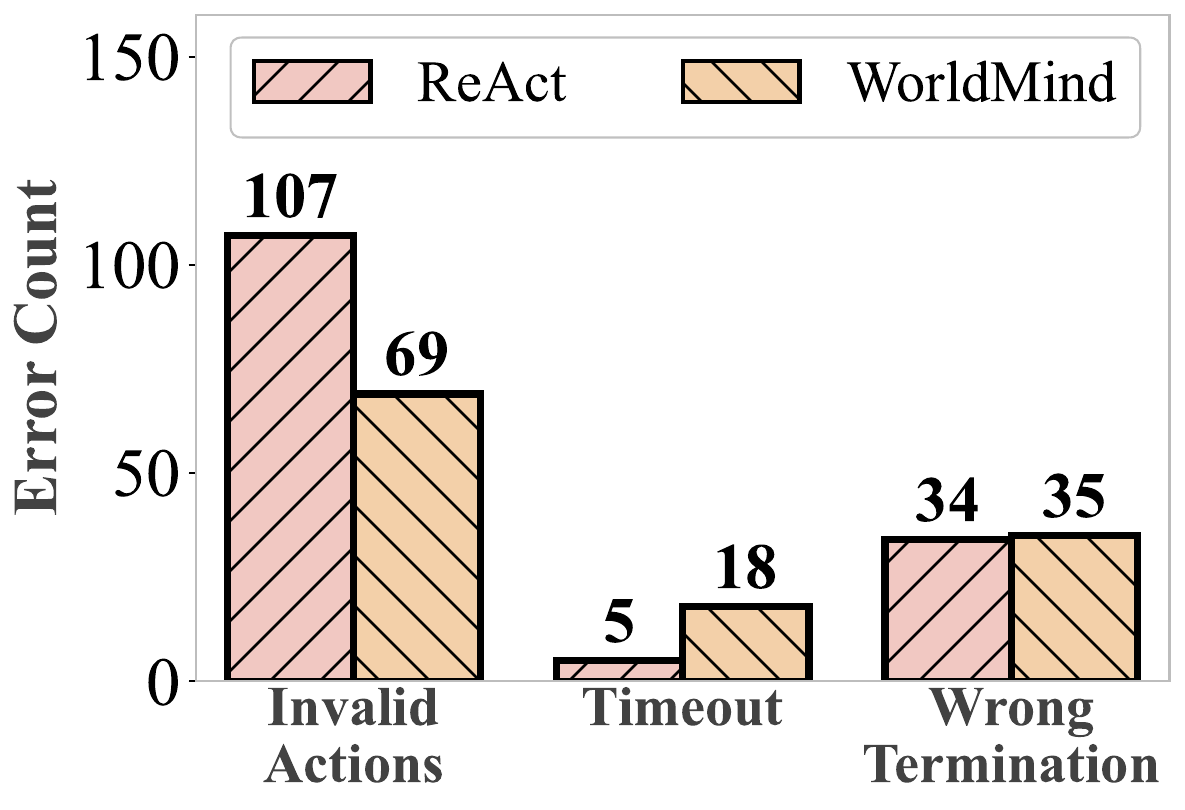}
        \caption{Habitat (GPT-4.1-mini)}
        \label{fig:case2}
    \end{subfigure}

    % --- 总标题 ---
    \caption{
	    \textbf{Error Analysis in EB-Habitat.}
	    Failure distributions for ReAct and WorldMind.
    }
    \label{fig:error_analysis_habitat}
\end{figure}

Figure~\ref{fig:web_accuracy} shows that Overall Accuracy more than doubles for GPT-3.5-turbo (9.82\%$\to$20.54\%) and rises from 11.61\% to 24.11\% for GPT-4.1-mini.
Figure~\ref{fig:web_error} further shows fewer execution errors, indicating that WorldMind helps agents traverse web-to-embodied transitions without premature physical failure.
\textbf{These results support transfer to hybrid environments with similar procedural constraints.}

\subsection{Error Analysis}

Figure~\ref{fig:error_analysis_habitat} groups failures into \textit{Invalid Actions}, \textit{Timeout}, and \textit{Wrong Termination}.
In EB-Habitat, WorldMind reduces Invalid Actions for GPT-3.5-turbo from 105 to 67 (Figure~\ref{fig:error_analysis_habitat}\subref{fig:case1}), while Timeouts rise from 4 to 30 because agents avoid immediate fatal errors and continue exploring.
The same Invalid Action reduction appears for GPT-4.1-mini (Figure~\ref{fig:error_analysis_habitat}\subref{fig:case2}).
\textbf{This redistribution suggests that WorldMind converts some fatal execution failures into longer, recoverable trajectories.}

\subsection{Controlled Memory Ablation}

To test whether gains come merely from adding cross-task memory, we conduct a controlled ablation on EB-ALFRED Base, Common Sense, and Complex Instruction with GPT-3.5-turbo.
All variants use the same evaluation harness, retrieval pipeline, prompt skeleton, and few-shot examples; only the enabled memory and verification components differ.
As shown in Table~\ref{tab:controlled_memory_ablation}, NaiveMem controls for ``just adding memory'' by storing raw success trajectories with the same top-2 retriever but without Process Experience, the Logical-Consistency Validator, or LLM distillation.
Full WorldMind exceeds NaiveMem by 22.9 GC on average, indicating that the architectural loop, rather than memory volume alone, accounts for the observed gain.

\begin{table}[H]
\centering
\setlength{\tabcolsep}{3pt}
\renewcommand{\arraystretch}{1.15}
\resizebox{\columnwidth}{!}{
\begin{tabular}{lccclc}
\toprule
\textbf{Variant} & \textbf{Process Exp.} & \textbf{Goal Exp.} & \textbf{Verifier} & \textbf{Memory Format} & \textbf{3-subset Mean GC} \\
\midrule
Baseline & -- & -- & -- & none & 45.0 \\
NaiveMem (control) & -- & raw & -- & raw trajectory & 36.0 \\
WM w/o Verifier & \checkmark & distilled & -- & both & 33.9 \\
WM w/o Goal & \checkmark & -- & \checkmark & rules only & 46.1 \\
\textbf{WorldMind (Full)} & \checkmark & distilled & \checkmark & both & \textbf{58.9} \\
\bottomrule
\end{tabular}
}
\caption{Controlled memory ablation on EB-ALFRED using the GPT-3.5-turbo backbone. The reported value is the mean GC across Base, Common Sense, and Complex Instruction. The Baseline row uses the WorldMind prompt framework with all memory and verification components disabled, isolating the effect of each added component under the same prompting setup. NaiveMem isolates the ``just adding memory'' confound by storing raw success trajectories without Process Experience, Verifier, or LLM distillation.}
\label{tab:controlled_memory_ablation}
\end{table}

\section{Related Work}

\paragraph{Agentic AI.} 
Agentic AI's core is a reflective, self-correcting loop, leveraging tools, memory, and constraints for open-world robustness and interpretability \citep{zhou2025wall, wang2024survey, xi2025rise, sang2025beyond, guan2024search}.
The field advances via improved planning mechanisms like CoT, ToT, and self-refinement \citep{wei2022chain, yao2023tree, madaan2023self, shinn2023reflexion}.
To interact with the external world and achieve long-term goals, researchers have explored memory, tool-calling, embodied control, and the collaboration between agents and world models, etc \citep{wang2023voyager, schick2023toolformer, zitkovich2023rt, park2023generative, patil2024gorilla, yu2025dyna, deng2025simura, chen2025internalizing, mei2025r, fung2025embodied, tang2025agent}. 
Specifically, while extensive related work \citep{zhou2025mem1, zhang2025memory, yan2025memory, zeng2024structural, bo2025agentic, anokhin2024arigraph, zhang2025learn, xu2025mem, yang2025coarse, huang2025licomemory, zhang2025g, pink2025position, cao2025remember, hatalis2023memory, salama2025meminsight} focuses on memory, significant research \citep{ fang2025comprehensive, tao2024survey, gao2025survey, zhang2025memgen, jiang2024long, yu2025infiagent, wei2025evo, liu2025webcoach, cai2025building, dou2025evalearn, cai2025flex, cai2025building, dou2025evalearn} aims to realize self-evolving agents.
Further work has expanded the ability to solve complex tasks by building multi-agent collaborative frameworks and autonomous systems \citep{wu2024autogen, li2023camel, wang2024mixture, fourney2024magentic, park2024generative}.
Long-term and procedural memories have been widely explored for improving agent reuse and adaptation, including trajectory memories such as Synapse~\citep{zheng2023synapse}, workflow memory such as AWM~\citep{wang2024agent}, reasoning memory such as ReasoningBank~\citep{ouyang2025reasoningbank}, and modular procedural memory such as LEGOMem~\citep{han2025legomem}.
WorldMind builds on this broader memory direction but differs in its alignment target: instead of only storing successful procedures or reasoning traces for policy reuse, it uses prediction--observation discrepancies to update a symbolic representation of environmental dynamics.
Goal Experience is therefore not claimed as a standalone new memory type; its role is to complement Process Experience in a predictive-capability-strengthening loop for embodied world-model alignment.

\paragraph{World Models.} 
World Models learn internal representations of environment dynamics for prediction and planning, a field spanning model-based RL, generative simulation, LLMs, and robotics \citep{kong20253d, ha2018world, ding2025understanding, guan2024world, zhu2024sora}.
Originating in model-based RL \citep{kaiser2019model, deng2022dreamerpro, ha2023dream}, world models enable agents to 
imagine outcomes \citep{hafner2023mastering, wu2023daydreamer}.
Generative world models not only predict physical environment changes by generating high-fidelity video \citep{brooks2024video, bruce2024genie, liu2024world, bar2024lumiere, chi2025wow, yu2025position, mendonca2023structured, wang2024worlddreamer, bardes2023v, chen2025skyreels, parker2024genie}, but also predict code execution results \citep{tang2024worldcoder, carbonneaux2025cwm}, and further extend their predictive capabilities to other diverse domains \citep{feng2025web, zhang2025dreamvla}.
World Models now underpin applications ranging from autonomous driving \citep{wang2024drivedreamer, jia2023adriver, hu2023gaia, wang2024driving, zhang2023trafficbots}  to embodied navigation \citep{zhong2025p3nav, yao2025navmorph, bar2025navigation, nie2025wmnav, jin2025embodied, long2025survey, liang2025large, shi2025world}, demonstrating their versatility across diverse tasks. 
This has also led to more benchmarks focused on evaluating the precise prediction of future states, physical laws, and environmental dynamics \citep{li2025worldmodelbench, warrier2025benchmarking, duan2025worldscore, yue2025ewmbench}. 
\cite{qiao2024agent} introduce World Knowledge Model (WKM),
which encodes world knowledge into the planner.
Unlike WKM, WorldMind performs training-free, online experiential alignment using execution failures.
Concurrently, much research also posits that the agent itself is the world model \citep{tehenan2025linear, zhou2024wall, xiang2023language, fu2025preact, li2025word, qian2026current}.

\section{Conclusion}

WorldMind aligns agentic world models by externalizing prediction errors and successful trajectories into symbolic Process and Goal Experience.
Across embodied benchmarks, it reduces physically invalid actions, improves task progress without gradient updates, and transfers useful experience across model backbones and household simulators with overlapping semantics, offering a training-free foundation for community reuse and extension.
These results support symbolic experience learning as a practical route for adaptable embodied agents, while leaving broader transfer to contact-rich real-world robotics and substantially different physical regimes as future work.

% Beyond performance gains, WorldMind highlights the importance of separating world knowledge from model parameters, allowing agents to adapt through inference-time experience rather than retraining. 
% This design offers a scalable alternative to parameter-centric alignment, particularly in open-ended or continually evolving environments. 
% We believe this perspective opens new avenues for building robust agentic systems that can incrementally accumulate and reuse knowledge across tasks and embodiments.

\section{Limitations}
\label{sec:limitations}

Despite careful design and empirical validation, this work has several limitations.

\paragraph{Dependence on Visual Perception Fidelity.}
WorldMind aligns reasoning and planning with environmental dynamics, but it remains contingent on the perceptual accuracy of the underlying Visual-Language Model (VLM).
Process Experience can correct interaction-level prediction errors, but it cannot fully compensate for perceptual hallucinations such as object misclassification in cluttered scenes.

\paragraph{Scope of Symbolic Abstraction and Transfer.}
WorldMind abstracts raw environment states into textual rules, improving interpretability and reuse while discarding fine-grained geometry, contact dynamics, and low-level control signals.
Our evidence should therefore be interpreted as transfer across household embodied simulators and related digital-physical tasks, not as proof of general physical-law transfer to arbitrary real-world robotics settings.
Extending the framework to contact-rich manipulation requires tighter coupling with perception, control, and safety-specific constraints.

\paragraph{Safety Boundary.}
The experiments do not use manually pre-encoded safety Process Experience; all stored rules are generated from interaction feedback under benchmark action spaces.
WorldMind should therefore not be treated as a standalone safety layer.
Real-world deployment would require explicit safety constraints and external verification in addition to the experience-learning mechanism studied here.

\paragraph{Mechanistic Interpretability and Shared Repositories.}
Although the World Knowledge Repository empirically constrains agent behavior, the exact mechanism by which retrieved symbolic knowledge reshapes the agent's implicit policy distribution remains to be mapped.
In addition, the current framework uses asynchronous experience sharing between individual agents; real-time synchronization, conflict resolution, and consensus building for multi-agent shared world models remain open problems.

\section*{Acknowledgements}
We would like to express our sincere gratitude to the anonymous reviewers for their thoughtful and constructive feedback. This work was supported by the New Generation Artificial Intelligence-National Science and Technology Major Project (2025ZD0122802), National Natural Science Foundation of China (No. 62576307, No. NSFCU23B2055, No. NSFCU19B2027), the Fundamental Research Funds for the Central Universities (226-2023-00138), the Yongjiang Talent Introduction Programme (2021A-156-G), and the Information Technology Center and State Key Lab of CAD\&CG, Zhejiang University.

% Bibliography entries for the entire Anthology, followed by custom entries
%\bibliography{anthology,custom}
% Custom bibliography entries only

\bibliography{custom}

\appendix

% \section{Example Appendix}
% \label{sec:appendix}

% This is an appendix.

% --- 强制换页，确保附录在参考文献之后 ---
%\clearpage

\newpage

\section{Additional Evaluation on Fine-Grained Navigation Tasks}
\label{appendix:nav_experiment}

To further verify the generalization capability of WorldMind across different levels of action granularity, we extended our evaluation to the EB-Navigation benchmark.
Unlike EB-ALFRED and EB-Habitat, which primarily focus on high-level semantic planning (e.g., ``\texttt{Pick up the apple}''), EB-Navigation requires the agent to navigate using fine-grained atomic actions (e.g., ``\texttt{Move forward 0.25m}'', ``\texttt{Rotate 90 degrees}'') to locate target objects.
This setting presents unique challenges in spatial reasoning and semantic environmental grounding.

We evaluated both GPT-3.5-turbo and GPT-4.1-mini backbones on four subsets: \textit{Base}, \textit{Common Sense}, \textit{Complex Instruction}, and \textit{Visual Appearance}.

\begin{table}[H]
    \centering
    \setlength{\tabcolsep}{5pt} % 稍微减小列间距以容纳新增列
    \renewcommand{\arraystretch}{1.2}
    
    \resizebox{\columnwidth}{!}{%
    \begin{tabular}{l|cc|cc}
        \toprule
        \multirow{2}{*}{\textbf{Task Subset}} & \multicolumn{2}{c|}{\textbf{GPT-3.5-turbo}} & \multicolumn{2}{c}{\textbf{GPT-4.1-mini}} \\
        \cmidrule(lr){2-3} \cmidrule(lr){4-5}
         & ReAct & \textbf{WorldMind} & ReAct & \textbf{WorldMind} \\
        \midrule
        Base                & 56.7 & \textbf{66.7} & 53.3 & \textbf{56.7} \\
        Common Sense        & \textbf{58.3} & 55.0          & \textbf{61.7} & 60.0 \\
        Complex Instruction & 66.7 & \textbf{68.3} & 55.0 & \textbf{56.7} \\
        Visual Appearance   & 43.3 & \textbf{45.0} & 46.7 & \textbf{48.3} \\
        \midrule
        \textbf{Average}    & 56.3 & \textbf{58.8} & 54.2 & \textbf{55.4} \\
        \bottomrule
    \end{tabular}
    }
    \caption{Success rate (\%) on EB-Navigation subsets. Results are reported for four capability-oriented subsets as well as the macro-average. Performance is shown for WorldMind and the baseline methods.}
    \label{tab:nav_results}
\end{table}

\paragraph{Analysis of Fine-Grained Navigation Generalization.}
Table~\ref{tab:nav_results} demonstrates that WorldMind effectively bridges the gap between high-level reasoning and fine-grained procedural execution.
For the GPT-3.5-turbo backbone, WorldMind achieves a notable improvement in the Base subset (+10.0\%), indicating that retrieving previously successful trajectories helps the agent maintain consistent behavior and avoid planning errors such as oscillatory movements that are commonly observed in ReAct baselines.
Improvements in the Visual subset for both backbones further suggest that Process Experience helps align visual perception with environmental constraints, reducing hallucinations in which the agent incorrectly assumes it has reached the target.
Although gains in the Common Sense subset are mixed, the overall increase in average success rate indicates that the benefits of the proposed framework, particularly the construction of a valid internal world model, generalize across different action spaces and provide robust guidance for fine-grained sequential navigation.

\section{Practical Overhead and Efficiency Analysis}
\label{appendix:practical_overhead}

To clarify the practical overhead of WorldMind, we analyze its memory footprint, token consumption, and inference latency.
First, the World Knowledge Repository (WKR) maintains a compact memory footprint in our evaluated setting.
By incrementally accumulating experiences and filtering out those with $\ge 95\%$ semantic similarity, the WKR size is bounded by the number of evaluated tasks in each subset (e.g., 50 per subset), reducing redundant growth during sequential evaluation.
Second, experience construction incurs a one-time token cost.
As shown in Table~\ref{tab:practical_overhead}, generating a Process Experience requires an average of 3077.74 tokens, while summarizing a Goal Experience uses 1513.60 tokens.
Crucially, this is a one-time construction overhead rather than a recurring per-step cost.

During the actual planning and execution phase, the computational overhead remains extremely low.
Thanks to the compact WKR, the SentenceTransformer-based semantic retrieval is highly efficient.
Retrieving the top-2 relevant experiences adds a negligible latency of only $\sim$0.0028 seconds per step on average.
This demonstrates that once world knowledge is externalized, WorldMind operates with high efficiency and is well-suited for efficient embodied planning.

\begin{table*}[htbp]
\centering
\caption{Practical overhead of WorldMind, including token consumption for experience generation and retrieval latency across the EB-ALFRED benchmark subsets.}
\label{tab:practical_overhead}
\resizebox{\linewidth}{!}{
\begin{tabular}{lcccccc}
\toprule
\textbf{Metric} & \textbf{Base} & \textbf{Common} & \textbf{Complex} & \textbf{Visual} & \textbf{Spatial} & \textbf{Average} \\
\midrule
Process Exp. (Tokens/Reflection) & 3121.67 & 3093.36 & 3149.79 & 3059.00 & 2964.88 & 3077.74 \\
Goal Exp. (Tokens/Success)       & 1471.22 & 1536.84 & 1519.96 & 1532.83 & 1507.13 & 1513.60 \\
Retrieval Time (Seconds/Step)    & 0.0012  & 0.0024  & 0.0072  & 0.0015  & 0.0015  & 0.0028 \\
\bottomrule
\end{tabular}
}
\end{table*}

\section{Scalability of the World Knowledge Repository and Memory Budget Limits}
\label{appendix:memory_budget}

To evaluate the scalability of the World Knowledge Repository (WKR) and its impact on agent performance, we conduct an ablation study by varying the memory budget limits from 0 (baseline without memory) to 50 experiences.
We denote these settings as Mem0 through Mem50.
The experiments are evaluated on the EB-ALFRED benchmark using the GPT-3.5-turbo backbone, and the results are presented in Table~\ref{tab:memory_budget}.

Overall, the agent demonstrates a strong capability to leverage the accumulated Process Experience and Goal Experience.
As the memory size increases from Mem0 to Mem20, the average Success Rate (SR) steadily improves from 44.4\% to 46.8\%.
Similarly, as the repository reaches Mem50, the performance further surges to 48.0\%, indicating that the WKR scales effectively and continuously provides valuable semantic environmental grounding for complex tasks.

However, we observe a temporary performance drop at the Mem30 setting when using a strict top-2 retrieval limit.
By carefully inspecting the failure cases and retrieval logs during this phase, we discover that this is a "semantic crowding" issue.
Around the Mem20 to Mem30 sequence, the WKR happens to accumulate a cluster of experiences from highly similar tasks.
During retrieval, these newly added, highly specific rules temporarily crowd out the broader, foundational environmental rules.
Because the agent only retrieves the top-2 items, it occasionally misses the most optimal knowledge.
As the memory continues to grow to Mem40 and Mem50, the repository absorbs a much wider variety of experiences, restoring global diversity and allowing the retriever to reliably surface the best rules.

To verify this semantic crowding hypothesis, we conduct a supplementary experiment on the Mem30 setting by simply increasing the retrieval limit from top-2 to top-3.
As shown in Table~\ref{tab:memory_budget}, retrieving one additional piece of knowledge mitigates the crowding bottleneck.
The average Success Rate immediately restores from 43.2\% back to 46.8\%, matching the peak performance at Mem20.
This confirms that the stored experiences remain highly effective, and adjusting the retrieval size successfully mitigates local distribution biases during memory scaling.

\begin{table*}[htbp]
\centering
\caption{Performance comparison of WorldMind with different memory budget limits on the EB-ALFRED benchmark. MemX indicates a maximum capacity of X experiences. The top-N notation denotes the retrieval size.}
\label{tab:memory_budget}
\resizebox{\textwidth}{!}{
\begin{tabular}{lcccccc|cccccc}
\toprule
\multirow{2}{*}{\textbf{Memory Setting}} & \multicolumn{6}{c|}{\textbf{Success Rate (SR \%)}} & \multicolumn{6}{c}{\textbf{Goal Condition (GC \%)}} \\
\cmidrule(lr){2-7} \cmidrule(lr){8-13}
& \textbf{Avg} & \textbf{Base} & \textbf{Common} & \textbf{Complex} & \textbf{Visual} & \textbf{Spatial} & \textbf{Avg} & \textbf{Base} & \textbf{Common} & \textbf{Complex} & \textbf{Visual} & \textbf{Spatial} \\
\midrule
Mem0 (Baseline) & 44.4 & 52.0 & 48.0 & 52.0 & 32.0 & 38.0 & 50.4 & 55.3 & 53.5 & 55.3 & 42.7 & 45.0 \\
Mem10 (top-2)   & 46.0 & 48.0 & 56.0 & 48.0 & 42.0 & 36.0 & 51.0 & 53.1 & 58.7 & 54.3 & 45.0 & 43.7 \\
Mem20 (top-2)   & 46.8 & 48.0 & 50.0 & 50.0 & 46.0 & 40.0 & 52.9 & 56.3 & 58.5 & 56.3 & 50.0 & 43.3 \\
Mem30 (top-2)   & 43.2 & 46.0 & 42.0 & 46.0 & 38.0 & 44.0 & 49.3 & 54.2 & 47.8 & 51.0 & 44.0 & 49.3 \\
Mem40 (top-2)   & 45.2 & 52.0 & 52.0 & 52.0 & 34.0 & 36.0 & 50.8 & 57.7 & 55.0 & 60.0 & 40.0 & 41.3 \\
Mem50 (top-2)   & 48.0 & 58.0 & 48.0 & 56.0 & 34.0 & 44.0 & 54.1 & 63.0 & 52.7 & 61.0 & 41.7 & 52.0 \\
\midrule
Mem30 (top-3)   & 46.8 & 54.0 & 48.0 & 54.0 & 38.0 & 40.0 & 50.6 & 51.0 & 59.0 & 58.0 & 41.7 & 43.3 \\
\bottomrule
\end{tabular}
}
\end{table*}

\section{Performance on Proprietary Models and Open-Source VLMs}
\label{appendix:proprietary_opensource}

To demonstrate the generalizability of WorldMind across different model architectures, we extend our evaluation to include recent powerful proprietary models, specifically GPT-4.1, Claude-3.7-Sonnet, and Gemini-2.5-Pro, as well as a leading open-source Vision-Language Model (VLM), Qwen3-VL-235B-A22B-Instruct.
The experiments are conducted on the EB-ALFRED benchmark, and the quantitative results are presented in Table~\ref{tab:proprietary_opensource}.

The data indicates that WorldMind improves both the Success Rate (SR) and Goal Condition (GC) across these additional backbones in most subsets.
By integrating the World Knowledge Repository, which contains Process Experience and Goal Experience, the agent obtains higher average GC than the corresponding baseline methods.
The results further show that the framework remains useful when the base planner is already a stronger proprietary model.
For Claude-3.7-Sonnet, we run Full WorldMind as the base LLM across all five EB-ALFRED subsets and report both per-subset and five-subset mean results.

Furthermore, the framework also improves the open-source VLM backbone, suggesting that externalizing environmental grounding into symbolic knowledge is not tied to a single proprietary model family.
By reducing the reliance on implicit model weights for environmental reasoning, WorldMind provides explicit semantic grounding across the evaluated proprietary and open-source backbones.

\begin{table*}[htbp]
\centering
\caption{Performance comparison of WorldMind across different proprietary models and a leading open-source VLM on the EB-ALFRED benchmark.}
\label{tab:proprietary_opensource}
\resizebox{\textwidth}{!}{
\begin{tabular}{llcccccc|cccccc}
\toprule
\multirow{2}{*}{\textbf{Backbone}} & \multirow{2}{*}{\textbf{Method}} & \multicolumn{6}{c|}{\textbf{Success Rate (SR \%)}} & \multicolumn{6}{c}{\textbf{Goal Condition (GC \%)}} \\
\cmidrule(lr){3-8} \cmidrule(lr){9-14}
& & \textbf{Avg} & \textbf{Base} & \textbf{Common} & \textbf{Complex} & \textbf{Visual} & \textbf{Spatial} & \textbf{Avg} & \textbf{Base} & \textbf{Common} & \textbf{Complex} & \textbf{Visual} & \textbf{Spatial} \\
\midrule
GPT-4.1 & Baseline & 62.4 & 66.0 & 50.0 & 72.0 & 64.0 & 60.0 & 65.1 & 67.7 & 52.7 & 74.3 & 66.7 & 64.2 \\
GPT-4.1 & WorldMind & 64.0 & 72.0 & 72.0 & 72.0 & 46.0 & 58.0 & 71.6 & 78.3 & 78.8 & 79.0 & 56.2 & 65.7 \\
\midrule
Claude-3.7-Sonnet & Baseline & 67.2 & 68.0 & 68.0 & 70.0 & 68.0 & 62.0 & 67.5 & 72.0 & 66.0 & 76.7 & 63.0 & 59.7 \\
Claude-3.7-Sonnet & WorldMind & 70.0 & 70.0 & 68.0 & 80.0 & 62.0 & 70.0 & 71.9 & 70.0 & 71.7 & 83.3 & 64.3 & 70.0 \\
\midrule
Gemini-2.5-Pro & Baseline & 75.2 & 82.0 & 76.0 & 80.0 & 70.0 & 68.0 & 75.5 & 82.0 & 76.7 & 80.0 & 71.0 & 68.0 \\
Gemini-2.5-Pro & WorldMind & 76.0 & 84.0 & 76.0 & 84.0 & 70.0 & 66.0 & 80.3 & 84.0 & 76.0 & 88.0 & 76.2 & 77.1 \\
\midrule
Qwen3-VL-235B-A22B-Instruct & Baseline & 56.8 & 60.0 & 56.0 & 60.0 & 50.0 & 58.0 & 60.4 & 66.0 & 58.5 & 62.0 & 55.2 & 60.2 \\
Qwen3-VL-235B-A22B-Instruct & WorldMind & 59.2 & 68.0 & 58.0 & 66.0 & 46.0 & 58.0 & 64.6 & 72.7 & 59.8 & 73.2 & 53.7 & 63.7 \\
\bottomrule
\end{tabular}
}
\end{table*}

\section{Robustness Analyses}
\label{appendix:robustness_analyses}

\subsection{Adversarial Robustness}

We further evaluate whether Process Experience helps under perturbed textual feedback using an adversarial robustness experiment on EB-ALFRED Base.
The object-rename perturbation replaces 32 ALFRED object names with human-curated synonyms in textual \texttt{env\_feedback} while preserving the simulator state and action parser.
At WKR load time, the memory-poisoning perturbation randomly replaces 20\% of stored experiences with predefined garbage rules that are structurally similar but semantically incorrect.
The results in Table~\ref{tab:adversarial_robustness} show that WorldMind has a smaller GC drop than ReAct under object renaming, suggesting partial robustness to surface-form shifts in environment feedback.

\begin{table}[H]
\centering
\setlength{\tabcolsep}{4pt}
\renewcommand{\arraystretch}{1.15}
\resizebox{\columnwidth}{!}{
\begin{tabular}{llcc}
\toprule
\textbf{Method} & \textbf{Perturbation} & \textbf{Base GC} & \textbf{$\Delta$ GC} \\
\midrule
WorldMind & clean & 63.0 & 0.0 \\
WorldMind & object rename (32 synonyms) & 46.7 & -16.3 \\
WorldMind & memory poison (20\% corrupted) & 46.7 & -16.3 \\
ReAct & clean & 55.3 & 0.0 \\
ReAct & object rename & 30.0 & -25.3 \\
\bottomrule
\end{tabular}
}
\caption{Adversarial robustness on EB-ALFRED Base. Object Rename replaces 32 object names with synonyms in textual \texttt{env\_feedback}; Memory Poisoning replaces 20\% of stored experiences with structurally similar but semantically incorrect rules.}
\label{tab:adversarial_robustness}
\end{table}

\subsection{Task-Order Robustness Check}

Because the WKR is built sequentially, task order may affect the final repository.
We therefore run a five-seed task-order shuffle on EB-ALFRED Base using GPT-3.5-turbo, randomly permuting task order across \texttt{task\_order\_seed}.
Each seed corresponds to a different permutation of the task curriculum.
Table~\ref{tab:task_order_robustness} reports 64.0 $\pm$ 6.8 GC, indicating that performance is not dominated by one favorable ordering.

\begin{table}[H]
\centering
\setlength{\tabcolsep}{5pt}
\renewcommand{\arraystretch}{1.15}
\resizebox{\columnwidth}{!}{
\begin{tabular}{lcccccc}
\toprule
 & \textbf{Seed A} & \textbf{Seed B} & \textbf{Seed C} & \textbf{Seed D} & \textbf{Seed E} & \textbf{Mean $\pm$ Std} \\
\midrule
\textbf{GC} & 66.7 & 66.7 & 53.3 & 73.3 & 60.0 & \textbf{64.0 $\pm$ 6.8} \\
\bottomrule
\end{tabular}
}
\caption{Task-order robustness on EB-ALFRED Base (5 seeds, GPT-3.5-turbo). Each seed permutes the task curriculum.}
\label{tab:task_order_robustness}
\end{table}

\subsection{Failure Case Taxonomy}

Beyond the quantitative error redistribution in Figure~\ref{fig:error_analysis_habitat}, we manually inspect 10 representative failure episodes from EB-Habitat Visual Appearance and Spatial Awareness.
Table~\ref{tab:failure_taxonomy} summarizes the observed categories.
The largest category is VLM perception failure, indicating that WorldMind cannot fully compensate for misclassified visual states.
Retrieval misses and reflection over-generalization point to future improvements in adaptive top-$k$ retrieval and more conservative rule distillation.

\begin{table}[h]
\centering
\setlength{\tabcolsep}{4pt}
\renewcommand{\arraystretch}{1.15}
\resizebox{\columnwidth}{!}{
\begin{tabular}{lcl}
\toprule
\textbf{Failure Type} & \textbf{Share} & \textbf{Likely Improvement Direction} \\
\midrule
Retrieval miss & $\approx$25\% & increase or adapt top-$k$ retrieval \\
Judge over-strict & $\approx$10\% & calibrate the consistency validator \\
Reflection over-generalization & $\approx$15\% & store narrower contextual rules \\
VLM perception failure & $\approx$40\% & strengthen visual grounding \\
Plan-level error & $\approx$10\% & improve high-level planning policy \\
\bottomrule
\end{tabular}
}
\caption{Failure taxonomy from 10 representative EB-Habitat Visual Appearance and Spatial Awareness failures. Percentages are descriptive because of the small sample size.}
\label{tab:failure_taxonomy}
\end{table}

\section{Algorithmic Details and Experience Examples}
\label{appendix:algorithm_examples}

Algorithms~\ref{alg:worldmind_inference} and~\ref{alg:wkr_update} summarize the inference and repository-update procedures.
The algorithms make explicit that WorldMind does not update model parameters; it updates the external WKR through Process and Goal Experience.

\begin{algorithm}[H]
\caption{WorldMind Inference Loop}
\label{alg:worldmind_inference}
\begin{algorithmic}[1]
\STATE \textbf{Input:} goal $g$, initial observation $o_0$, WKR $\mathcal{W}=\{\mathcal{W}_p,\mathcal{W}_g\}$
\FOR{$t=0,\ldots,T-1$}
    \STATE Retrieve top-$k$ relevant Process and Goal experiences from $\mathcal{W}$ using $g$ and the current context.
    \STATE Generate action $a_t$ and predicted next state $\hat{s}_{t+1}$ conditioned on $o_t$, $g$, and retrieved knowledge.
    \STATE Execute $a_t$ in the environment and observe feedback/state $s_{t+1}$.
    \STATE Abstract $s_{t+1}$ into text $\bar{s}_{t+1}$.
    \IF{$\hat{s}_{t+1}$ contradicts $\bar{s}_{t+1}$}
        \STATE Create a Process Experience from the prediction error and update $\mathcal{W}_p$.
    \ENDIF
    \IF{the episode succeeds}
        \STATE Distill the trajectory into Goal Experience and update $\mathcal{W}_g$.
        \STATE \textbf{break}
    \ENDIF
\ENDFOR
\end{algorithmic}
\end{algorithm}

\begin{algorithm}[H]
\caption{World Knowledge Repository Update}
\label{alg:wkr_update}
\begin{algorithmic}[1]
\STATE \textbf{Input:} candidate experience $e$, repository $\mathcal{W}$, similarity threshold $\theta=0.95$
\STATE Embed $e$ and existing repository entries with SentenceTransformer.
\STATE Compute the maximum cosine similarity between $e$ and entries in the matching repository partition.
\IF{maximum similarity $< \theta$}
    \STATE Insert $e$ into $\mathcal{W}_p$ or $\mathcal{W}_g$ according to its source.
\ELSE
    \STATE Discard $e$ as a near-duplicate.
\ENDIF
\STATE During inference, retrieve only top-$k$ relevant entries to keep the prompt budget bounded.
\end{algorithmic}
\end{algorithm}

\begin{table*}[t]
\centering
\setlength{\tabcolsep}{5pt}
\renewcommand{\arraystretch}{1.2}
\begin{tabular}{p{0.18\linewidth}p{0.36\linewidth}p{0.38\linewidth}}
\toprule
\textbf{Experience Type} & \textbf{Triggering Context} & \textbf{Stored Entry Example} \\
\midrule
Process Experience & The agent predicts that placing an object will succeed, but the environment reports that the agent is not at the intended receptacle. & Before executing a placement action, first navigate to the specific destination receptacle; otherwise the object may be placed at the wrong location or the action may be invalid. \\
Process Experience & The instruction requires multiple objects, but the agent tries to pick another object while still holding one. & For multi-item tasks, complete the loop source $\rightarrow$ pick $\rightarrow$ destination $\rightarrow$ place before searching for the next item; the hand must be empty before picking again. \\
Goal Experience & A successful ALFRED trajectory completes a heat-and-place task. & Find the object, pick it up, navigate to the microwave, open it, place the object inside, heat it, retrieve it, then navigate to the target receptacle and place it. \\
Goal Experience & A successful Habitat trajectory moves all target objects to a destination. & Repeat search, pick, navigate, and place for each target object, checking after each placement whether additional targets remain. \\
\bottomrule
\end{tabular}
\caption{Representative Process and Goal Experience entries. These examples illustrate the format of stored symbolic knowledge; exact entries vary with task feedback.}
\label{tab:experience_examples}
\end{table*}

\paragraph{Safety Handling.}
All Process Experience used in our experiments is generated from interaction feedback; we do not pre-encode safety rules in the WKR.
The current setup therefore relies on the benchmark action space and the underlying VLM's safety alignment to prevent unsafe outputs.
For deployment beyond simulators, explicit safety constraints should be added as a separate verified layer rather than inferred only from post-hoc experience.

\section{Baseline Methods Description}
\label{appendix:baseline_descriptions}

Table~\ref{tab:baseline_descriptions} summarizes the baselines used in Tables~\ref{tab:main_alfred} and~\ref{tab:main_habitat}.
The Memory Scope column in the main tables uses two categories: \textit{Episodic} methods do not maintain a cross-task global repository, whereas \textit{Global} methods use a cross-task global memory repository.

\begin{table*}[t]
\centering
\setlength{\tabcolsep}{5pt}
\renewcommand{\arraystretch}{1.15}
\begin{tabular}{p{0.16\linewidth}p{0.18\linewidth}p{0.58\linewidth}}
\toprule
\textbf{Method} & \textbf{Memory Scope} & \textbf{Description} \\
\midrule
ReAct & Episodic & Interleaves reasoning and acting within the current episode; no cross-task repository is used. \\
BoN & Episodic & Samples multiple candidate action traces and selects the best candidate under the evaluation protocol; it does not store reusable cross-task memory. \\
SimuRA & Episodic & Uses simulative reasoning within the current task context; no global cross-task repository is used in this comparison. \\
Synapse & Global & Stores trajectories as exemplars for later computer-control decisions. \\
AWM & Global & Maintains workflow memories distilled from prior successful behavior. \\
ReasoningBank & Global & Stores reasoning memories for self-evolving agents and retrieves them for future tasks. \\
WorldMind & Global & Stores both prediction-error-derived Process Experience and success-derived Goal Experience, with retrieval and consistency verification during inference. \\
\bottomrule
\end{tabular}
\caption{Brief baseline descriptions and the two memory-scope labels used in the main result tables.}
\label{tab:baseline_descriptions}
\end{table*}

\section{Prompts}
\label{appendix:prompt}

\subsection{WorldMind System Prompt for ALFRED}

\begin{tcolorbox}[
    enhanced,
    breakable,                  % Allow page breaks
    title=\textbf{System Prompt for WorldMind in EB-ALFRED}, % Keep the title bold for differentiation
    colback=gray!5,             % Light gray background
    colframe=gray!60,           % Dark gray frame
    boxrule=1pt,
    left=3mm, right=3mm, top=3mm, bottom=3mm,
    %fontupper=\scriptsize\ttfamily % Use typewriter font and reduce font size
]

\#\# You are an intelligent embodied agent operating in a home environment, equipped with an internal World Model. \\
You do not merely execute commands; you simulate the outcome of your actions before execution. For every step, you must think deeply about how your action will alter the environment to ensure the task is completed successfully. Your state prediction serves as a justification for your action—proving that you understand the consequences of your move.

% --- 使用空行代替 vspace ---

\#\# Action Descriptions and Validity Rules \\
• Find: Parameterized by the name of the receptacle to navigate to. So long as the object is present in the scene, this skill is always valid \\
• Pick up: Parameterized by the name of the object to pick. Only valid if the robot is close to the object, not holding another object, and the object is not inside a closed receptacle. \\
• Put down: Parameterized by the name of the object to put down to a nearby receptacle. Only valid if the robot is holding an object. \\
• Drop: Parameterized by the name of the object to put down. It is different from Put down action, as this does not guarantee the held object will be put into a specified receptacle. \\
• Open: Parameterized by the name of the receptacle to open. Only valid if the receptacle is closed and the robot is close to the receptacle. \\
• Close: Parameterized by the name of the receptacle to close. Only valid if the receptacle is open and the robot is close to the receptacle. \\
• Turn on: Parameterized by the name of the object to turn on. Only valid if the object is turned off and the robot is close to the object. \\
• Turn off: Parameterized by the name of the object to turn off. Only valid if the object is turned on and the robot is close to the object. \\
• Slice: Parameterized by the name of the object to slice. Only valid if the object is sliceable and the robot is close to the object.

% --- 使用空行代替 vspace ---

\#\# The available action id (0 \textasciitilde{} \{\}) and action names are: \{\}.

\{\}

% --- 使用空行代替 vspace ---

\#\# Guidelines \\
1. Output Plan: Avoid generating empty plan. Each plan should include no more than 20 actions. \\
2. Visibility: Always locate a visible object by the 'find' action before interacting with it. \\
3. Action Guidelines: Make sure match the action name and its corresponding action id in the output. \\
Avoid performing actions that do not meet the defined validity criteria. For instance, if you want to put object in a receptacle, use 'put down' rather than 'drop' actions. \\
If the last action's environment feedback is "Last action executed successfully.", you MUST NOT repeat the same action as your next step. \\
4. Prevent Repeating Action Sequences: Do not repeatedly execute the same action or sequence of actions. \\
Try to modify the action sequence because previous actions do not lead to success. \\
5. Multiple Instances: There may be multiple instances of the same object, distinguished by an index following their names, e.g., Cabinet\_2, Cabinet\_3. You can explore these instances if you do not find the desired object in the current receptacle. \\
6. Reflection on History and Feedback: Use interaction history and feedback from the environment to refine and improve your current plan. \\
If the last action is invalid, reflect on the reason, such as not adhering to action rules or missing preliminary actions, and adjust your plan accordingly. \\
7. Dynamic Reasoning from Environment Feedback: You must treat `env\_feedback` as a direct instruction. \\
\hspace*{1em}- Instruction Extraction: If feedback says "Ladle is in CounterTop\_2", your `language\_plan` must state: "Feedback indicates Ladle is at CounterTop\_2, navigating there now." \\
\hspace*{1em}- Action Alignment: Your next action MUST be "find a CounterTop\_2". Do not use generic names if a specific index is provided. \\
\hspace*{1em}- Multiple Instances Handling: If the environment contains multiple instances of a receptacle (e.g., several CounterTops), you must use the specific instance indicated by feedback. Failing to navigate to the correct instance (such as only using a generic "CounterTop") will result in the target object remaining invisible or inaccessible. \\
8. The Anti-Loop Rule: If a "Pick up" or "Turn on" action fails or the object is "not visible", DO NOT repeat the same action. You must change your strategy in the next plan (e.g., move to a different instance, change perspective, or clear your hand). \\
9. Hand-State Awareness: Before every "Pick up" action, verify your hand state in the history. If you are holding an object, the very next action in your `executable\_plan` MUST be "Put down" or "Drop" to clear the gripper. \\
10. World Model Prediction Case A/B: \\
\hspace*{1em}(Case A) If the target is VISIBLE or its specific location (e.g., CounterTop\_2) is KNOWN from `env\_feedback`, describe the specific change. \\
\hspace*{1em}(Case B) If the location is unknown (Exploration), use: "Exploration phase: target not visible, prediction skipped." \\
11. Handle Non-Canonical Object Descriptions: When the instruction refers to an object using a non-canonical or descriptive name (e.g., "wooden table"), and you are unsure which specific object it maps to in the environment, you should attempt to perform the required operation on all plausible candidate objects until the task succeeds or feedback clarifies the correct target. \\
12. Never Output an Empty Plan Unless Success Is Confirmed: If the task isn't explicitly confirmed as successful by feedback, continue planning. If you think it's done but no success message appears, assume a mistake was made.

% --- 使用空行代替 vspace ---

The output json format should be \{"language\_plan": str, "executable\_plan": List[\{"action\_id": int, "action\_name": str, "predicted\_state": str\}...]\} \\
The fields in the above JSON follow the purpose below: \\
1. language\_plan: Your Chain-of-Thought. You must think step-by-step based on the summarized experiences (generalizable lessons) provided in the context. Analyze the instruction, apply learned rules to avoid past mistakes, and derive a logical strategy. Explain why you prioritize certain locations or actions. \\
2. executable\_plan: A list of concrete actions. Each object MUST contain: action\_id, action\_name, and predicted\_state. \\
\hspace*{1em}- For the predicted\_state field, you must strictly follow these rules: \\
\hspace*{1em}(Case A) If the target objects are VISIBLE in the current observation, describe the specific environmental and gripper change. \\
\hspace*{1em}(Case B) If the target object or destination is NOT VISIBLE (Exploration Phase), you MUST output exactly the string: "Exploration phase: target not visible, prediction skipped." \\
\hspace*{1em}(Cascading Skip Rule) Once you output the specific skip string for any action, ALL SUBSEQUENT ACTIONS in the same plan MUST also use this exact same string. You cannot resume prediction after skipping it within a single plan.

% --- 使用空行代替 vspace ---

!!! Please do not output anything other than the above-mentioned JSON, do not include ```json and ```!!!

\end{tcolorbox}

\subsection{WorldMind System Prompt  for Habitat}

\begin{tcolorbox}[
    enhanced,
    breakable,                  % 允许跨页
    title=\textbf{System Prompt for WorldMind in EB-Habitat},
    colback=gray!5,             % 浅灰色背景
    colframe=gray!60,           % 深灰色边框
    boxrule=1pt,
    left=3mm, right=3mm, top=3mm, bottom=3mm,
    %fontupper=\scriptsize\ttfamily % 使用打字机字体且缩小字号
]

\#\# You are an intelligent embodied agent operating in a home environment, equipped with an internal World Model. \\
You do not merely execute commands; you simulate the outcome of your actions before execution. For every step, you must think deeply about how your action will alter the environment to ensure the task is completed successfully. Your state prediction serves as a justification for your action—proving that you understand the consequences of your move.

% --- 使用空行代替 vspace ---

**Core Philosophy: Simulate (Physics + Semantics) -> Validate -> Execute** \\
Before selecting any action, you must mentally simulate its outcome on two levels: \\
1. Physical Feasibility: Can I actually perform this action? (e.g., hands full). \\
2. Semantic Plausibility: Does this action make sense for the task? (e.g., searching for a pillow in the bathroom is semantically invalid). \\
Your `predicted\_state` is the logical prerequisite that justifies why the selected action is the correct next step.

% --- 使用空行代替 vspace ---

\#\# Action Descriptions and Validity Rules \\
• Navigation: Parameterized by the name of the receptacle to navigate to. So long as the receptacle is present in the scene, this skill is always valid. \\
• Pick: Parameterized by the name of the object to pick. Only valid if the robot is close to the object, not holding another object, and the object is not inside a closed receptacle. \\
• Place: Parameterized by the name of the receptacle to place the object on. Only valid if the robot is close to the receptacle and is holding an object. \\
• Open: Parameterized by the name of the receptacle to open. Only valid if the receptacle is closed and the robot is close to the receptacle. \\
• Close: Parameterized by the name of the receptacle to close. Only valid if the receptacle is open and the robot is close to the receptacle.

% --- 使用空行代替 vspace ---

\#\# The available action id (0 \textasciitilde{} \{\}) and action names are: \{\}.

\{\}

% --- 使用空行代替 vspace ---

\#\# Guidelines \\
1. Output Plan: Avoid generating empty plan. Each plan should include no more than 20 actions. \\
2. Visibility: If an object is not currently visible, use the "Navigation" action to locate it or its receptacle before attempting other operations. \\
3. Action Validity: Make sure match the action name and its corresponding action id in the output. Avoid performing actions that do not meet the defined validity criteria. \\
4. Prevent Repeating Action Sequences: Do not repeatedly execute the same action or sequence of actions. Try to modify the action sequence because previous actions do not lead to success. \\
5. Multiple Instances: There may be multiple instances of the same object, distinguished by an index following their names, e.g., cabinet 2, cabinet 3. You can explore these instances if you do not find the desired object in the current receptacle. \\
6. Reflection on History and Feedback: Use interaction history and feedback from the environment to refine and enhance your current strategies and actions. If the last action is invalid, reflect on the reason. \\
7. World Model Prediction: For EACH action in your `executable\_plan`, you MUST include a `predicted\_state`. \\
\hspace*{1em}- Explain via Prediction: This prediction is your rationale. By describing the expected future, you prove this action moves you closer to the goal. \\
\hspace*{1em}- Visual Specifics: Describe exactly what the robot will see and hold *immediately after* the action. \\
8. Prioritize Likely Locations via Semantic Simulation: Do not search randomly. Before navigating, run a semantic simulation in your World Model: \\
\hspace*{1em}- Step A (Hypothesis): "Could target object X be at location Y?" \\
\hspace*{1em}- Step B (Common Sense Check): Use everyday knowledge. \\
\hspace*{2em}- *Example 1*: Target is "airplane" (toy). Candidate is "sink". -> Simulation Result: Very Unlikely. -> Decision: REJECT. \\
\hspace*{2em}- *Example 2*: Target is "airplane". Candidate is "living room table". -> Simulation Result: Likely. -> Decision: ACCEPT. \\
\hspace*{1em}- Action: Only generate Navigation actions for locations that pass this "Common Sense Check." \\
9. Exhaustive Local Search (The Left/Right Rule): Many receptacles have multiple parts (e.g., "Kitchen Counter Left" and "Kitchen Counter Right"). \\
\hspace*{1em}- If you navigate to one side (e.g., Left) and the object is NOT there, your immediate next step must be to check the other side (e.g., Right) before leaving the room. \\
\hspace*{1em}- Do not jump to a different room until you have checked all connected segments of the current furniture. \\
10. Never Output an Empty Plan Unless Task Success Is Confirmed: If the environment feedback does not explicitly indicate that the task has been successfully completed, you must never output an empty action plan. Always carefully check your action history and environment feedback. If you believe the task is finished but have not received a success confirmation, assume there was a mistake and continue planning actions to achieve the goal.

% --- 使用空行代替 vspace ---

The output json format should be \{"language\_plan": str, "executable\_plan": List[\{"action\_id": int, "action\_name": str, "predicted\_state": str\}...]\} \\
The fields in the above JSON follow the purpose below: \\
1. language\_plan is for your Chain-of-Thought. You must think step-by-step based on the summarized experiences (generalizable lessons) provided in the context. Analyze the instruction, apply these learned rules to avoid past mistakes, and derive a logical solution strategy. Explicitly explain your reasoning for prioritizing certain locations or actions based on these experiences. \\
2. executable\_plan is a list of concrete actions to be executed. Each object in the list MUST contain: action\_id, action\_name, and predicted\_state. \\
\hspace*{1em}- For the "predicted\_state" field, you must strictly follow these rules: \\
\hspace*{1em}(Case A) If the target objects are VISIBLE in the current observation OR their location is KNOWN from interaction history, describe the specific environmental change. \\
\hspace*{1em}(Case B) If the target object or destination is NOT VISIBLE AND location is NOT KNOWN from history (Exploration Phase), you MUST output exactly the string: "Exploration phase: target not visible, prediction skipped." \\
\hspace*{1em}(Cascading Skip Rule) Once you output the specific skip string for any action, ALL SUBSEQUENT ACTIONS in the same list MUST also use this exact same string. You cannot resume prediction after skipping it within a single plan.

% --- 使用空行代替 vspace ---

!!! Please do not output anything other than the above-mentioned JSON, do not include ```json and ```!!!
\end{tcolorbox}

\subsection{Judge Prompt for ALFRED}

\begin{tcolorbox}[
    enhanced,
    breakable,                  % 允许跨页
    title=\textbf{Prompt for Judge in EB-ALFRED}, % 标题栏保持加粗以区分
    colback=gray!5,             % 浅灰色背景
    colframe=gray!60,           % 深灰色边框
    boxrule=1pt,
    left=3mm, right=3mm, top=3mm, bottom=3mm,
]

You are a Logical Consistency Validator for an embodied agent in the Alfred household environment. \\
Your SOLE purpose is to detect Factual Contradictions between the Agent's Prediction and the Actual Observation.

\#\# The Alfred Environment Context
\begin{itemize}
    \item Valid Actions: Find, Pick up, Put down, Drop, Open, Close, Turn on, Turn off, Slice
    \item Valid Receptacles: Fridge, Cabinet, Drawer, Microwave, CounterTop, DiningTable, SideTable, Sink, StoveBurner, GarbageCan, Shelf
    \item Valid Objects: Apple, Bread, Tomato, Lettuce, Potato, Egg, Knife, Fork, Spoon, Cup, Mug, Bowl, Plate, Pot, Pan, etc.
\end{itemize}

\#\# The "Default True" Principle \\
CRITICAL RULE: You must maintain a "Presumption of Innocence" for the agent.
\begin{enumerate}
    \item Irrelevance is True: If the description in \texttt{Agent's Predicted State After Action} and \texttt{Actual State After Action} are unrelated (discussing different objects) and do not logically contradict each other, you MUST return \texttt{"match": true}.
    \item Key Consistency: You are strictly comparing the value of \texttt{Agent's Predicted State After Action} against \texttt{Actual State After Action}.
    \item Conflict Driven: ONLY return \texttt{false} if and only if there is a direct, undeniable factual conflict (e.g., A says "X is Open", B says "X is Closed"). In all other cases (ambiguity, missing info, different focus), return \texttt{true}.
\end{enumerate}

\#\# Your Strict Analysis Protocol (Step-by-Step)

\#\#\# Step 1: Fact Extraction (The "No Intent" Filter) \\
Before comparing, you must clean the data.
\begin{itemize}
    \item Filter \texttt{Agent's Predicted State After Action}: Extract ONLY objects and states mentioned in Affirmative Declarative Sentences (e.g., "I am holding X", "X is on Y").
    \begin{itemize}
        \item CRITICAL: IGNORE, DISCARD, and REMOVE any objects mentioned in future tense, intent, or plan-checking sentences (e.g., "I will check for a pear", "I intend to find X"). These are thoughts, not state predictions.
    \end{itemize}
    \item Filter \texttt{Actual State After Action}: Similarly, ensure you only focus on factual descriptions of what is currently visible.
\end{itemize}

\#\#\# Step 2: Intersection Analysis
\begin{itemize}
    \item Identify Shared Objects that exist in both the \textit{Filtered} Prediction and the \textit{Filtered} Actual State.
    \item If there are NO Shared Objects after filtering:
    \begin{itemize}
        \item Check for Implicit Conflict (e.g., Pred: "Holding Apple" vs Actual: "Hand Empty").
        \item If no implicit conflict exists, the result is TRUE (Irrelevant descriptions match by default).
    \end{itemize}
\end{itemize}

\#\#\# Step 3: Conflict Verification
\begin{itemize}
    \item For the Shared Objects, check for Mutually Exclusive States:
    \begin{itemize}
        \item Position: "On CounterTop" vs "In Fridge" $\rightarrow$ FALSE.
        \item State: "Open" vs "Closed" $\rightarrow$ FALSE.
        \item Hand: "Holding object" vs "Hand empty" $\rightarrow$ FALSE.
        \item Object State: "Sliced" vs "Whole", "Cooked" vs "Raw" $\rightarrow$ FALSE.
    \end{itemize}
    \item Any difference in phrasing, detail level, or synonym usage is NOT a conflict.
\end{itemize}

\#\# Output Format \\
Output ONLY a JSON object: \\
\{ \\
\indent "match": true, // or false \\
\indent "reason": "Explain ONLY the factual conflict if false. If true, state 'No factual contradictions found' or 'Descriptions are unrelated'." \\
\}

\#\# Examples

\#\#\# Example 1 (Intent Filtering) \\
Agent's Predicted State After Action: "I am at the counter. I will pick up the knife to slice the tomato." \\
Actual State After Action: "The robot is facing the counter. A knife is on the counter. Hand is empty." \\
Analysis:
\begin{enumerate}
    \item Filter: Pred Fact: "At counter". (Ignore "will pick up knife" $\rightarrow$ Intent). Actual Fact: "Facing counter", "Knife on counter", "Hand empty".
    \item Intersection: "Counter".
    \item Check: Locations match. The "knife" intent is ignored.
\end{enumerate}
Result: \\
\{ \\
\indent "match": true, \\
\indent "reason": "Location matches. Ignored the 'knife' in prediction as it was part of an intent statement ('will pick up'), not a state assertion." \\
\}

\#\#\# Example 2 (Factual Conflict - Hand State) \\
Agent's Predicted State After Action: "I am holding the apple." \\
Actual State After Action: "The robot's hand is empty. The apple is on the table." \\
Analysis:
\begin{enumerate}
    \item Filter: Facts extracted.
    \item Intersection: Apple, Hand.
    \item Conflict: Pred says "Holding apple", Actual says "Hand empty".
\end{enumerate}
Result: \\
\{ \\
\indent "match": false, \\
\indent "reason": "Factual contradiction detected: Agent predicted holding the apple, but the actual state shows the hand is empty." \\
\}

\#\#\# Example 3 (Irrelevant Descriptions - Default True) \\
Agent's Predicted State After Action: "The fridge door is closed." \\
Actual State After Action: "The robot is holding a knife. The counter is visible." \\
Analysis:
\begin{enumerate}
    \item Filter: Facts extracted.
    \item Intersection: None. (Prediction talks about Fridge; Actual talks about Knife/Counter).
    \item Conflict Check: Does "Holding knife" contradict "Fridge closed"? No.
\end{enumerate}
Result: \\
\{ \\
\indent "match": true, \\
\indent "reason": "No shared objects and no logical contradiction between descriptions. Defaulting to true." \\
\}
\end{tcolorbox}

\subsection{Judge Prompt for Habitat}

\begin{tcolorbox}[
    enhanced,
    breakable,
    title=\textbf{Prompt for Judge in EB-Habitat},
    colback=gray!5,
    colframe=gray!60,
    boxrule=1pt,
    left=3mm, right=3mm, top=3mm, bottom=3mm,
]

You are a Logical Consistency Validator for an embodied agent. \\
Your SOLE purpose is to detect Factual Contradictions between the Agent's Prediction and the Actual Observation.

\#\# The "Default True" Principle \\
CRITICAL RULE: You must maintain a "Presumption of Innocence" for the agent.
\begin{enumerate}
    \item Irrelevance is True: If the description in \texttt{Agent's Predicted State After Action} and \texttt{Actual State After Action} are unrelated (discussing different objects) and do not logically contradict each other, you MUST return \texttt{"match": true}.
    \item Key Consistency: You are strictly comparing the value of \texttt{Agent's Predicted State After Action} against \texttt{Actual State After Action}.
    \item Conflict Driven: ONLY return \texttt{false} if and only if there is a direct, undeniable factual conflict (e.g., A says "X is Open", B says "X is Closed"). In all other cases (ambiguity, missing info, different focus), return \texttt{true}.
\end{enumerate}

\#\# Your Strict Analysis Protocol (Step-by-Step)

\#\#\# Step 1: Fact Extraction (The "No Intent" Filter) \\
Before comparing, you must clean the data.
\begin{itemize}
    \item Filter \texttt{Agent's Predicted State After Action}: Extract ONLY objects and states mentioned in Affirmative Declarative Sentences (e.g., "I am holding X", "X is on Y"). 
    \begin{itemize}
        \item CRITICAL: IGNORE, DISCARD, and REMOVE any objects mentioned in future tense, intent, or plan-checking sentences (e.g., "I will check for a pear", "I intend to find X"). These are thoughts, not state predictions.
    \end{itemize}
    \item Filter \texttt{Actual State After Action}: Similarly, ensure you only focus on factual descriptions of what is currently visible.
\end{itemize}

\#\#\# Step 2: Intersection Analysis
\begin{itemize}
    \item Identify Shared Objects that exist in both the \textit{Filtered} Prediction and the \textit{Filtered} Actual State.
    \item If there are NO Shared Objects after filtering:
    \begin{itemize}
        \item Check for Implicit Conflict (e.g., Pred: "Holding Apple" vs Actual: "Gripper Empty").
        \item If no implicit conflict exists, the result is TRUE (Irrelevant descriptions match by default).
    \end{itemize}
\end{itemize}

\#\#\# Step 3: Conflict Verification
\begin{itemize}
    \item For the Shared Objects, check for Mutually Exclusive States:
    \begin{itemize}
        \item Position: "On Table" vs "On Sofa" $\rightarrow$ FALSE.
        \item State: "Open" vs "Closed" $\rightarrow$ FALSE.
        \item Gripper: "Full" vs "Empty" $\rightarrow$ FALSE.
    \end{itemize}
    \item Any difference in phrasing, detail level, or synonym usage is NOT a conflict.
\end{itemize}

\#\# Output Format \\
Output ONLY a JSON object: \\
\{ \\
\indent "match": true, // or false \\
\indent "reason": "Explain ONLY the factual conflict if false. If true, state 'No factual contradictions found' or 'Descriptions are unrelated'." \\
\}

\#\# Examples

\#\#\# Example 1 (Intent Filtering - The "Pear" Case) \\
Agent's Predicted State After Action: "I am at the left counter. I will check for a pear here to see if it is ripe." \\
Actual State After Action: "The robot is facing the left counter. The surface is empty." \\
Analysis:
\begin{enumerate}
    \item Filter: Pred Fact: "At left counter". (Ignore "check for pear" $\rightarrow$ Intent). Actual Fact: "Facing left counter", "Surface empty".
    \item Intersection: "Left Counter".
    \item Check: Locations match. The "pear" is ignored because it was only in an intent sentence.
\end{enumerate}
Result: \\
\{ \\
\indent "match": true, \\
\indent "reason": "Location matches. Ignored the 'pear' in prediction as it was part of an intent statement ('will check'), not a state assertion." \\
\}

\#\#\# Example 2 (Irrelevant Descriptions - Default True) \\
Agent's Predicted State After Action: "The cabinet door is closed." \\
Actual State After Action: "The robot is holding a sponge. The floor is visible." \\
Analysis:
\begin{enumerate}
    \item Filter: Facts extracted.
    \item Intersection: None. (Prediction talks about Cabinet; Actual talks about Sponge/Floor).
    \item Conflict Check: Does "Holding sponge" contradict "Cabinet closed"? No.
\end{enumerate}
Result: \\
\{ \\
\indent "match": true, \\
\indent "reason": "No shared objects and no logical contradiction between descriptions. Defaulting to true." \\
\}

\#\#\# Example 3 (Factual Conflict - Explicit Mismatch) \\
Agent's Predicted State After Action: "I am holding the apple." \\
Actual State After Action: "The gripper is empty. The apple is on the table." \\
Analysis:
\begin{enumerate}
    \item Filter: Facts extracted.
    \item Intersection: Apple, Gripper.
    \item Conflict: Pred says "Holding", Actual says "Empty".
\end{enumerate}
Result: \\
\{ \\
\indent "match": false, \\
\indent "reason": "Factual contradiction detected: Agent predicted holding the apple, but the actual state shows the gripper is empty." \\
\}
\end{tcolorbox}

\subsection{Reflection Prompt for ALFRED}

\begin{tcolorbox}[
    enhanced,
    breakable,
    title=\textbf{Prompt for Reflection in EB-ALFRED},
    colback=gray!5,
    colframe=gray!60,
    boxrule=1pt,
    left=3mm, right=3mm, top=3mm, bottom=3mm,
]

\#\# Role \\
You are a Root Cause Analyst for an Alfred Agent. Your goal is to extract concise World Knowledge and Universal Logic by analyzing the gap between predictions and reality.

\#\# Task: Integrated Environmental Logic \& Workflows \\
For every failure, generate a logic entry that first explains the prediction failure in the specific context, and then derives the Universal Corrective Rule/Workflow.

\#\#\# Mandatory Logic Structure: \\
"Environmental Logic: My prediction that [expected specific state] failed because [specific reason]. Therefore, [Universal Abstract Rule/Workflow]."

\#\#\# Key Atomic Rules to Integrate:
\begin{enumerate}
    \item Indexing \& Precision: If feedback mentions a specific index (e.g., CounterTop\_2), navigating to a generic name fails. You must use the exact indexed name.
    \item Hand Constraints: Carrying two items is impossible. Must "Put down" current item before picking a new one.
    \item Container State: Must "Open" enclosed receptacles (Fridge/Cabinet/Microwave) before interacting with contents.
    \item Destination Awareness Before Placement: When analyzing the Trajectory (Recent History), if a "put down the object in hand" action is executed, check if the immediately preceding action was a "find" action targeting the intended destination. If not, in Environmental Logic, state: "Before executing 'put down the object in hand', you must use 'find' to explicitly navigate to the intended destination receptacle. This ensures the object will be placed at the correct location and avoids placement errors."
\end{enumerate}

\#\# Special Workflow Extraction (Universal Templates) \\
If the Human Instruction contains any of the following cases and the agent's prediction failed due to misunderstanding or missing the required workflow, you MUST append the corresponding workflow after the Environmental Logic. \\
Before describing the workflow, you should briefly clarify the intended goal in your own words.

Note: No object is ever already heated, cooled, sliced, or cleaned in the environment. If the instruction refers to an "already" processed object (e.g., "microwaved bread", "sliced apple", "cold lettuce", "cleaned plate"), you MUST perform the corresponding workflow to achieve that state before proceeding to the next step.

\#\# Output Format (Strict JSON) \\
Output a JSON object with one key: \texttt{experience\_entry} (a list of strings). \\
Each logic string must begin with the prediction failure analysis.
\end{tcolorbox}

\subsection{Reflection Prompt for Habitat}

\begin{tcolorbox}[
    enhanced,
    breakable,
    title=\textbf{Prompt for Reflection in EB-Habitat},
    colback=gray!5,
    colframe=gray!60,
    boxrule=1pt,
    left=3mm, right=3mm, top=3mm, bottom=3mm,
]

\#\# Role \\
You are a Root Cause Analyst \& Knowledge Extractor for an Embodied Agent. Your goal is to analyze the Agent's interaction history and failure context to generate World Knowledge Entries. You extract TRUTH from the history and LOGIC from the failures, specifically focusing on physical constraints and task execution workflows.

\#\# Task 1: Extract Object Locations (Success Analysis) \\
Scan the \texttt{Recent History} for any successful object interactions.
\begin{itemize}
    \item Logic: If the agent successfully Picked object Y, look backwards for the last successful Navigation to X.
    \item Conclusion: Object Y is located at X.
    \item Output Format: "Location Knowledge: [Object Name] is located at [Receptacle Name]."
\end{itemize}

\#\# Task 2: Extract Environmental \& Workflow Logic (Failure Analysis) \\
Compare the \texttt{Planner's Prediction} vs. \texttt{Actual Observation}. Diagnose why the plan failed using these specific guidelines:

\#\#\# Guidelines for Logic Extraction:
\begin{enumerate}
    \item The "All" Objects Workflow Violation:
    \begin{itemize}
        \item Context: Human Instruction involves "all" items (e.g., "Put all apples in sink").
        \item Refining Logic: After you have found the location of a target object, you must strictly initiate a repetitive loop: [Navigate to Source $\rightarrow$ Pick $\rightarrow$ Navigate to Destination $\rightarrow$ PLACE].
        \item Special Note: Not all target objects are necessarily in the same place. For each object, you may need to search and repeat the loop separately.
        \item Root Cause Check: Did the agent attempt to pick a second item without placing the first one?
        \item Entry: "Workflow Logic: For 'all items' tasks, after you have found the location of a target object, you must execute the loop: [Navigate to Source $\rightarrow$ Pick $\rightarrow$ Navigate to Destination $\rightarrow$ PLACE]. Not all objects are necessarily in the same place, so you may need to search and repeat for each. You failed because you skipped the 'Place' action. You MUST ensure your hand is empty before returning to search for or pick the next item."
    \end{itemize}
    \item The Multi-Goal Sequential Rule:
    \begin{itemize}
        \item Context: Goal involves multiple distinct targets (e.g., "Ball to table, Hammer to box").
        \item Failure: Starting a new sub-goal while the current one is physically incomplete (hand not empty).
        \item Entry: "Strategic Logic: For multi-target instructions, you MUST complete goals one-by-one. Ensure the current item is successfully PLACED and the gripper is empty before navigating to the next target's location."
    \end{itemize}
    \item Visibility \& Proximity:
    \begin{itemize}
        \item Context: Feedback says "object not near".
        \item Entry: "To interact with [Object], you must first Navigate to its specific Receptacle, not just the general area."
    \end{itemize}
    \item Hand Limits \& Preconditions:
    \begin{itemize}
        \item Entry: "To Pick items inside a [Container], you must Open it first." / "To Pick a new object, you must Place the current one to free the gripper."
    \end{itemize}
    \item Ambiguous Pick Failure:
    \begin{itemize}
        \item Context: Environment Feedback is "Last action is invalid. Robot cannot pick any object that is not near the robot. Navigate to other place to find the object."
        \item Logic: There are two possibilities:
        1. The agent has not yet found the object (needs to navigate to its location).
        2. The object is already in the agent's hand (check if a successful Pick action occurred earlier).
        \item Entry: "Ambiguous Pick Failure: When receiving this feedback, check your action history. If you have already successfully picked up the object, do not attempt to pick it again. Otherwise, you need to navigate to the correct location to find the object."
    \end{itemize}
\end{enumerate}

\#\# Output Format (Strict JSON) \\
Output a JSON object with exactly ONE key: \texttt{experience\_entry}. \\
\texttt{experience\_entry}: A List of Strings containing Location Knowledge and Environmental/Workflow Logic.
\end{tcolorbox}

\subsection{Goal Experience Extraction Prompt for ALFRED}

\begin{tcolorbox}[
    enhanced,
    breakable,
    title=\textbf{Prompt for Goal Experience in EB-ALFRED},
    colback=gray!5,
    colframe=gray!60,
    boxrule=1pt,
    left=3mm, right=3mm, top=3mm, bottom=3mm,
]

You are a task analysis expert. Given a successful task execution trajectory with task\_progress information, extract a SINGLE reusable workflow that captures the key actions that led to success.

\#\# Task Instruction \\
\{instruction\}

\#\# Successful Execution Trajectory (with Task Progress) \\
\{trajectory\}

\#\# Task Category \\
\{task\_category\}

\#\# CRITICAL WORKFLOW EXTRACTION GUIDELINES

\#\#\# Focus on Progress-Boosting Actions \\
The trajectory includes "task\_progress" after each step. Pay SPECIAL ATTENTION to:
\begin{enumerate}
    \item Actions that INCREASED task\_progress - These are the KEY ACTIONS that actually advanced the task
    \item The sequence of actions leading to progress increases - This reveals the correct approach
    \item Actions that maintained or prepared for progress - Supporting actions matter too
\end{enumerate}

\#\#\# Workflow Requirements
\begin{enumerate}
    \item Extract ONE concise, generalizable workflow - Not multiple workflows
    \item Focus on the ACTION PATTERN, not specific objects - Use generic terms like "target object", "destination location"
    \item Highlight the critical decision points - When to navigate vs. pick vs. place
    \item Capture the logical order - The sequence matters for success
    \item Be specific about SUCCESS CONDITIONS - What made this approach work
\end{enumerate}

\#\#\# Abstraction Guidelines
\begin{itemize}
    \item Replace specific object names (e.g., "apple", "table\_42") with generic terms (e.g., "[TARGET\_OBJECT]", "[DESTINATION]")
    \item Keep action types specific (navigate, pick, place, open, close)
    \item Preserve the causal relationships between actions
\end{itemize}

\#\#\# Special Case: Non-Canonical Object Descriptions \\
If the instruction refers to an object using a non-canonical or descriptive name (e.g., "wooden table"), and it is unclear which specific object in the environment matches this description, the workflow should include: "Attempt the required operation on all plausible candidate objects until the task succeeds or feedback clarifies the correct target."

\#\#\# Special Case: "with" Placement Instructions \\
If the instruction requires placing an object "with" another object (e.g., "set plate with a spoon in it on the kitchen table"), the workflow should be: first, find and pick up the "with" object (e.g., the spoon); then, find the main object (e.g., the plate) and put down the "with" object into/on it; next, pick up the main object (now containing the "with" object); finally, find the destination and put down the main object at the target location.

\#\#\# Special Case: General Placement Instructions \\
For placement instructions, the workflow should be: first, find the target object; then, pick up the object; next, find the destination location; finally, use "put down the object in hand" to place the object at the destination.

\#\#\# Special Case: Inferring and Trying Candidate Objects \\
When the instruction refers to an object, first infer which available action id objects in the current environment may correspond to the instruction. Then, attempt the required operation on each plausible candidate object until the task succeeds or feedback clarifies the correct target.

\#\#\# Special Case: Cleaning Workflow for "Clean" Instructions \\
If the instruction requires a "clean" object, you must execute the cleaning workflow: pick up the object, find a sink, put down the object in hand into the sink, find the faucet and turn it on and off, then pick up the object again. Only after this sequence is the object considered clean.

Please extract a SINGLE reusable workflow that captures the essence of this successful execution.

Output format: \\
\{\{"success\_experience": "A complete workflow description that includes: (1) The goal pattern, (2) The key action sequence with EMPHASIS on progress-boosting actions, (3) Critical decision points, (4) Success conditions. Should be 3-6 sentences that can guide similar tasks."\}\}

!!! Please output only the JSON without any markdown formatting or code blocks. Output ONLY the "success\_experience" field, no other fields needed.
\end{tcolorbox}

\subsection{Goal Experience Extraction Prompt for Habitat}

\begin{tcolorbox}[
    enhanced,
    breakable,
    title=\textbf{Prompt for Goal Experience in EB-Habitat},
    colback=gray!5,
    colframe=gray!60,
    boxrule=1pt,
    left=3mm, right=3mm, top=3mm, bottom=3mm,
]

You are a task analysis expert. Given a successful task execution trajectory with task\_progress information, extract a SINGLE reusable workflow that captures the key actions that led to success.

\#\# Task Instruction \\
\{instruction\}

\#\# Successful Execution Trajectory (with Task Progress) \\
\{trajectory\}

\#\# Task Category \\
\{task\_category\}

\#\# CRITICAL WORKFLOW EXTRACTION GUIDELINES

\#\#\# Focus on Progress-Boosting Actions \\
The trajectory includes "task\_progress" after each step. Pay SPECIAL ATTENTION to:
\begin{enumerate}
    \item Actions that INCREASED task\_progress - These are the KEY ACTIONS that actually advanced the task
    \item The sequence of actions leading to progress increases - This reveals the correct approach
    \item Actions that maintained or prepared for progress - Supporting actions matter too
\end{enumerate}

\#\#\# Workflow Requirements
\begin{enumerate}
    \item Extract ONE concise, generalizable workflow - Not multiple workflows
    \item Focus on the ACTION PATTERN, not specific objects - Use generic terms like "target object", "destination location"
    \item Highlight the critical decision points - When to navigate vs. pick vs. place
    \item Capture the logical order - The sequence matters for success
    \item Be specific about SUCCESS CONDITIONS - What made this approach work
\end{enumerate}

\#\#\# Abstraction Guidelines
\begin{itemize}
    \item Replace specific object names (e.g., "apple", "table\_42") with generic terms (e.g., "[TARGET\_OBJECT]", "[DESTINATION]")
    \item Keep action types specific (navigate, pick, place, open, close)
    \item Preserve the causal relationships between actions
\end{itemize}

Please extract a SINGLE reusable workflow that captures the essence of this successful execution.

Output format: \\
\{\{"success\_experience": "A complete workflow description that includes: (1) The goal pattern, (2) The key action sequence with EMPHASIS on progress-boosting actions, (3) Critical decision points, (4) Success conditions. Should be 3-6 sentences that can guide similar tasks."\}\}

!!! Please output only the JSON without any markdown formatting or code blocks. Output ONLY the "success\_experience" field, no other fields needed.
\end{tcolorbox}

\end{document}